%% file: main.tex
\documentclass[10pt,journal,compsoc]{IEEEtran}

\ifCLASSOPTIONcompsoc
  \usepackage[nocompress]{cite}
\else
  \usepackage{cite}
\fi

\usepackage{times}
\usepackage{epsfig}
\usepackage{graphicx}
\usepackage{amsmath}
\usepackage{amssymb}
\usepackage{subfigure}
\usepackage{hyperref}
\usepackage{seqsplit}
\usepackage{url}
\usepackage{multirow}
\usepackage{threeparttable}
\usepackage{tablefootnote}
\usepackage{booktabs}
\usepackage{arydshln}
\usepackage{color}
\usepackage{xcolor}
\usepackage{tikz}
\usepackage{bm}
\usepackage{subfiles}
\usepackage{colortbl}
\newcommand{\PreserveBackslash}[1]{\let\temp=\\#1\let\\=\temp}
\newcolumntype{C}[1]{>{\PreserveBackslash\centering}p{#1}}

\ifCLASSOPTIONcompsoc
  \usepackage[nocompress]{cite}
\else
  \usepackage{cite}
\fi

\ifCLASSINFOpdf
\else
\fi

\hypersetup{
    colorlinks=true,
    linkcolor=blue,
    filecolor=magenta,      
    urlcolor=magenta,
}
\def\etal{\emph{et al}. }
\def\ie{\emph{i.e.} }
\def\eg{\emph{e.g.} }

\definecolor{Gray}{gray}{0.92}

\definecolor{codegreen}{rgb}{0.0,0.6,0.0}

\usepackage[ruled,vlined,linesnumbered]{algorithm2e}
\makeatletter
\newcommand{\algorithmfootnote}[2][\footnotesize]{%
  \let\old@algocf@finish\@algocf@finish% Store algorithm finish macro
  \def\@algocf@finish{\old@algocf@finish% Update finish macro to insert "footnote"
    \leavevmode\rlap{\begin{minipage}{\linewidth}
    #1#2
    \end{minipage}}%
  }%
}
\makeatother

\usepackage[accsupp]{axessibility}

\usepackage[capitalize]{cleveref}
\crefname{section}{Sec.}{Secs.}
\Crefname{section}{Section}{Sections}
\Crefname{table}{Table}{Tables}
\crefname{table}{Tab.}{Tabs.}

\newcommand{\myparagraph}[1]{{\vspace{.5em} \noindent \bf #1}}

\begin{document}

\newcommand{\modelname}{ByteTrackV2\xspace}

\title{\modelname: 2D and 3D Multi-Object Tracking by Associating Every Detection Box}

\author{Yifu Zhang,
        Xinggang Wang,
        Xiaoqing Ye,
        Wei Zhang,
        Jincheng Lu,
        Xiao Tan,
        Errui Ding,
        Peize Sun,
        Jingdong Wang
        
\IEEEcompsocitemizethanks{\IEEEcompsocthanksitem 
Corresponding author: Xinggang Wang. Email: xgwang@hust.edu.cn\\
\IEEEcompsocthanksitem Y. Zhang, X. Ye, W. Zhang, J. Lu, X. Tan, E. Ding, J. Wang are with Baidu Inc., China.\protect\\
\IEEEcompsocthanksitem X. Wang is with Huazhong University of Science and Technology, China.\protect\\
\IEEEcompsocthanksitem P. Sun is with The University of Hong Kong, China.\protect\\}
% E-mail: \{chnuwa, wezeng\}@microsoft.com
% \IEEEcompsocthanksitem Yifu Zhang, Xinggang Wang and Wenyu Liu are with Huazhong University of Science and Technology.\protect\\
% Email: \{yifuzhang, xgwang, liuwy\}@hust.edu.cn}% <-this % stops an unwanted space
}

% \markboth{Journal of \LaTeX\ Class Files,~Vol.~14, No.~8, August~2015}%
% {Shell \MakeLowercase{\textit{et al.}}: Bare Demo of IEEEtran.cls for Computer Society Journals}

\IEEEcompsoctitleabstractindextext{
\begin{abstract}
% Multi-object tracking (MOT) aims at estimating bounding boxes and identities of objects across video frames. 2D and 3D MOT tasks have been independently addressed as separate challenges by different research communities, which fails to fully leverage the commonalities of the two tasks. We propose \modelname to tackle the two tasks in a unified framework, including object detection, motion prediction, and detection-driven hierarchical data association. Detection boxes serve as the basis of the entire framework. We present a hierarchical data association strategy based on detection scores to mine the true objects in low-score detection boxes, which alleviates the problems of true object missing and fragmented trajectories. The generic data association strategy shows effectiveness under both 2D and 3D settings. In 3D scenarios, it is much easier for the tracker to predict object velocities in the world coordinate. We propose a complementary motion prediction strategy that incorporates the detected velocities with a Kalman filter to address the problem of abrupt motion and short-term disappearing.  \modelname leads the nuScenes 3D MOT leaderboard, ranking first in both camera (56.4\% AMOTA) and LiDAR (70.1\% AMOTA) modalities.  Furthermore, it is nonparametric and can be integrated with various detectors, facilitating the unification of 2D and 3D MOT tasks.

Multi-object tracking (MOT) aims at estimating bounding boxes and identities of objects across video frames. Detection boxes serve as the basis of both 2D and 3D MOT. The inevitable changing of detection scores leads to object missing after tracking. We propose a hierarchical data association strategy to mine the true objects in low-score detection boxes, which alleviates the problems of object missing and fragmented trajectories. The simple and generic data association strategy shows effectiveness under both 2D and 3D settings. In 3D scenarios, it is much easier for the tracker to predict object velocities in the world coordinate. We propose a complementary motion prediction strategy that incorporates the detected velocities with a Kalman filter to address the problem of abrupt motion and short-term disappearing. \modelname leads the nuScenes 3D MOT leaderboard in both camera (56.4\% AMOTA) and LiDAR (70.1\% AMOTA) modalities. Furthermore, it is nonparametric and can be integrated with various detectors, making it appealing in real applications. The source code is released at \url{https://github.com/ifzhang/ByteTrack-V2}.
 
\end{abstract}

}

\maketitle

\IEEEdisplaynotcompsoctitleabstractindextext

\IEEEpeerreviewmaketitle

%\IEEEraisesectionheading{\section{Introduction}}
\section{Introduction}
\subfile{1introduction}

\section{Related Work}
\subfile{2relatedwork}

\section{ByteTrackV2}
\subfile{3ByteTrack}

\section{Datasets and Metrics}
\subfile{5datasets}

\section{Experiments}
\subfile{6experiments}

\section{Conclusion}
\subfile{7conclusion}

% \ifCLASSOPTIONcompsoc
%   \section*{Acknowledgments}
% \else
%   \section*{Acknowledgment}
% \fi

\ifCLASSOPTIONcaptionsoff
  \newpage
\fi

\bibliographystyle{IEEEtran}      % basic style, author-year
\bibliography{egbib}   % name your BibTeX data base

\vfill

\end{document}

%% file: 1introduction.tex
\IEEEPARstart{T}{}his work addresses the problem of 2D and 3D multi-object tracking (MOT). Both 2D and 3D multi-object tracking \cite{milan2013continuous,bae2014robust,bewley2016simple,wojke2017simple,luo2018fast,baser2019fantrack,weng20203d} have been longstanding tasks in computer vision. The goal is to estimate the trajectories for objects of interest either in the 2D image plane or in the 3D world coordinates. Successfully solving the problem can benefit many applications such as autonomous driving and intelligent transportation.

\begin{figure}
\centering
\includegraphics[width=0.5\textwidth]{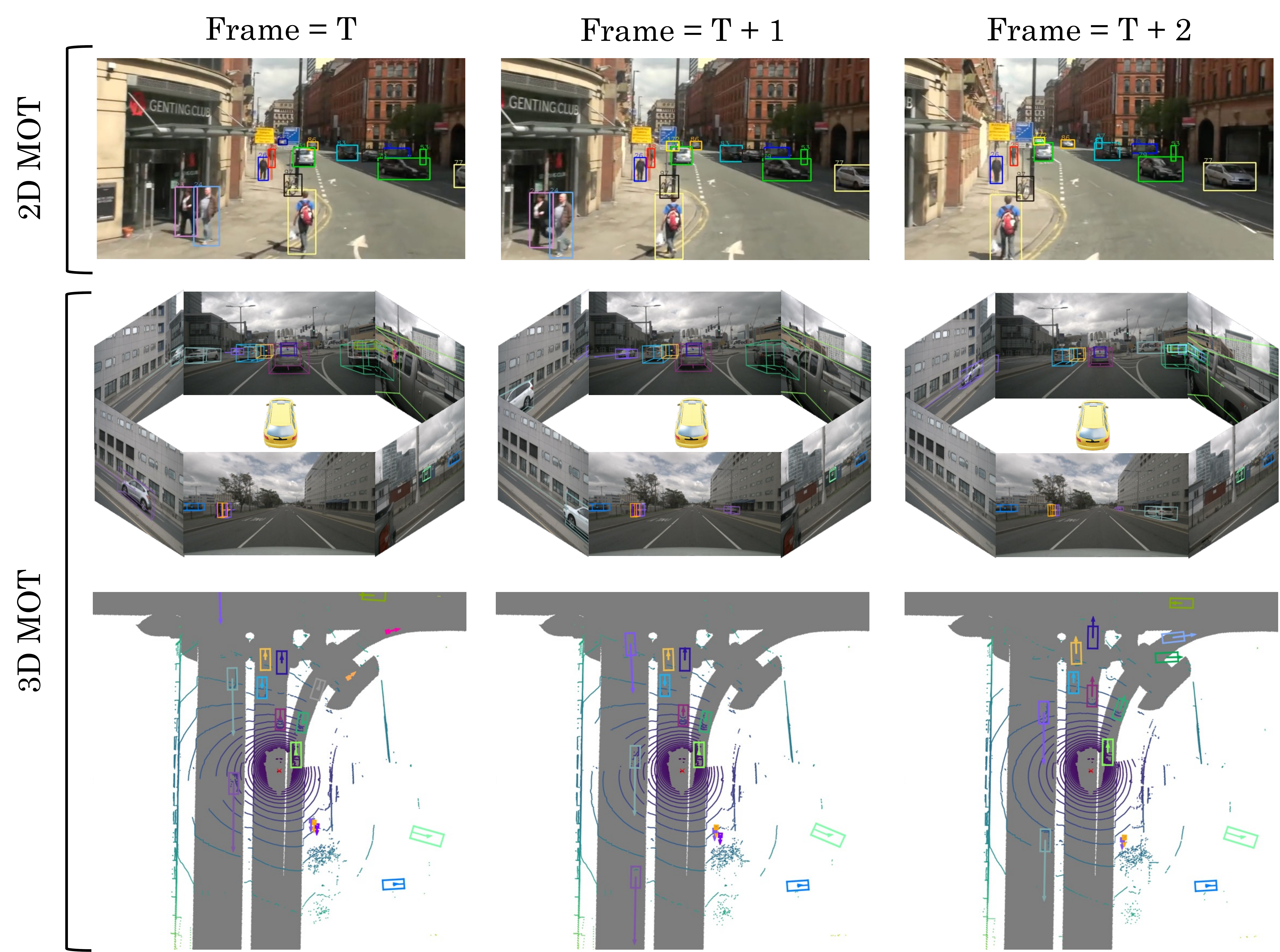}
\caption{Illustration of 2D multi-object tracking and 3D multi-object tracking. The first row shows the visualization of 2D MOT, which is performed on the image plane. The second row and the third row show the visualization of 3D MOT from the multi-view images and the Bird's Eye View (BEV) of the LiDAR point clouds, respectively. The same colors represent the same object identities.}
\label{fig:teasing_3d}
\vspace{-5mm}
\end{figure}

2D multi-object tracking and 3D multi-object tracking are inherently intertwined. Both tasks have to localize the objects and obtain object correspondence across different frames. However, they have been independently addressed by researchers from different areas because the input data come from different modalities. 2D MOT is carried out on the image plane and image information serves as an important cue for object correspondence. Appearance-based trackers \cite{bae2014robust,wojke2017simple,zhang2020fairmot,pang2021quasi} extract object appearance features from images and then compute the feature distance as correspondence. 3D MOT is usually performed in the world system which contains the depth information. It is easier to distinguish different objects by spatial similarity such as 3D Intersection over Union (IoU) \cite{weng20203d,weng2020ab3dmot} or point distance \cite{yin2021center}. Fig.~\ref{fig:teasing_3d} shows the visualization of 2D MOT and 3D MOT.

We solve the tasks of 2D and 3D MOT with three modules, \textit{i.e.}, detection, motion prediction, and data association. First, an object detector generates 2D / 3D detection boxes and scores. In the beginning frame, the detected objects are initialized as trajectories (or tracklets). Then, a motion predictor such as the Kalman filter \cite{kalman1960new} predicts the location of the tracklets in the following frame. Motion prediction is easy to realize on both image planes and the 3D world space. Finally, the detection boxes are associated with the predicted location of tracklets according to some spatial similarities. 

\begin{figure}
\centering
\includegraphics[width=0.45\textwidth]{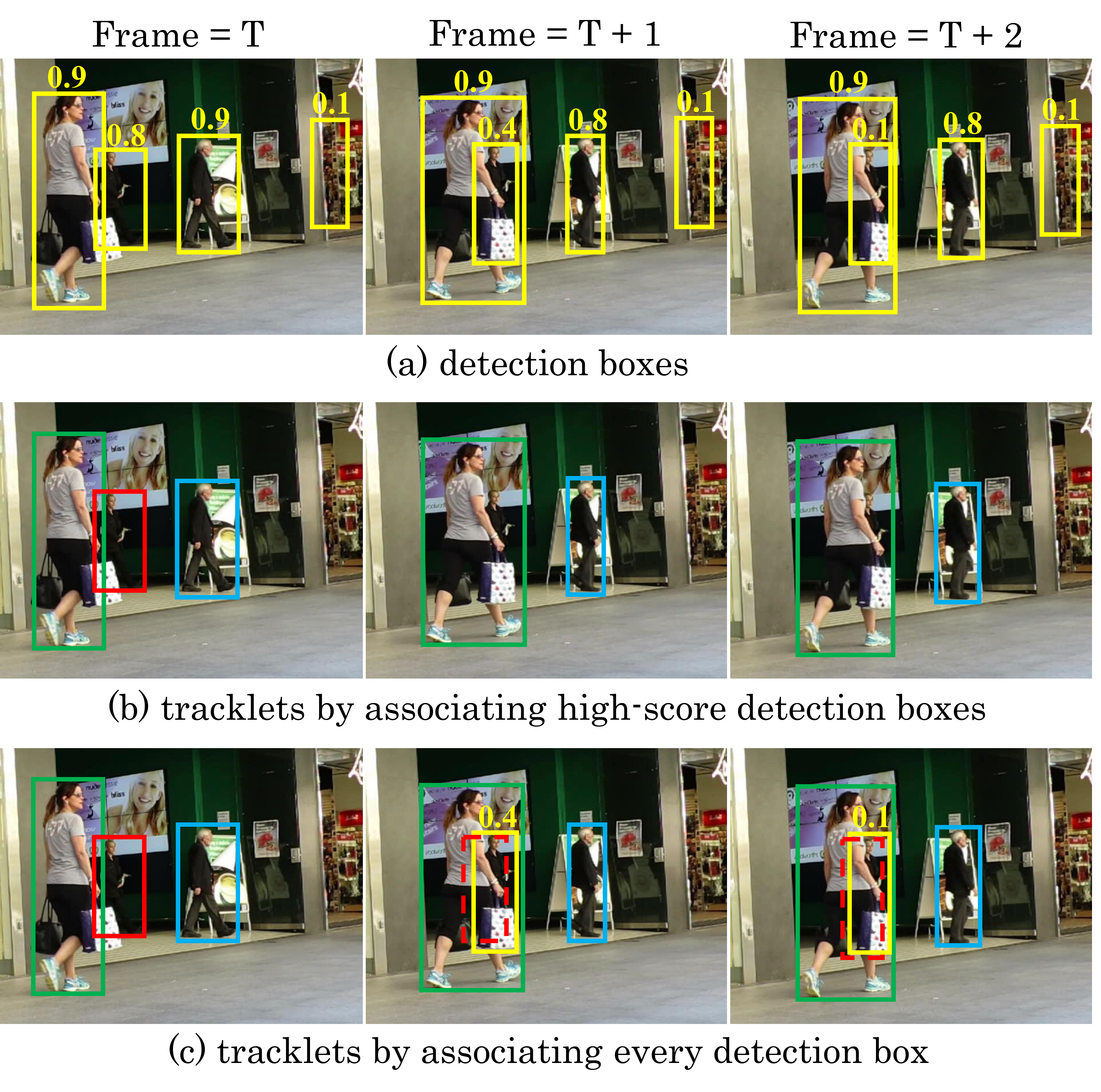}
\caption{Examples of our method that associates every detection box. (a) shows all the detection boxes with their scores. (b) shows the tracklets obtained by previous methods that associate detection boxes whose scores are higher than a threshold, \ie 0.5. The same box color represents the same identity. (c) shows the tracklets obtained by our method. The dashed boxes represent the predicted box of the previous tracklets using Kalman filter. The two low-score detection boxes are correctly matched to the previous tracklets based on the large IoU. The number colored in yellow denotes the score of the box.}
\label{fig:ass}
\vspace{-3mm}
\end{figure}

Detection is the basis of the entire MOT framework. Due to the complex scenarios in videos, detectors are prone to make imperfect predictions. High-score detection boxes usually contain more true positives than low-score ones. However, simply eliminating all low-score boxes is sub-optimal since low-score detection boxes sometimes indicate the existence of objects, \eg the occluded objects. Filtering out these objects causes irreversible errors for MOT and brings non-negligible missing detection and fragmented trajectories, as shown in Row (b) of Fig.~\ref{fig:ass}.

To address the problem of missing detection and fragmented trajectories caused by eliminating low-score boxes, we propose a detection-driven hierarchical data association strategy. It makes full use of detection boxes from high scores to low ones. We identify that the motion similarity between the detection boxes and tracklets provides a strong cue to distinguish the objects from the background in low-score detection boxes. We first associate the high-score detection boxes with the tracklets based on the motion similarity. Similar to \cite{bewley2016simple}, we adopt the Kalman filter to predict the location of the tracklets in the new frame. The similarity can be computed by the 2D or 3D IoU of the predicted box and the detection box. Then, we perform the second association between the unmatched tracklets and the low-score detection boxes using the same motion similarity to recover the true objects and remove the background. The association result is shown in line (c) of Fig.~\ref{fig:ass}.

In 3D MOT especially the driving scenarios, object abrupt motions and short-term disappearing caused by occlusion or blur bring about identity switches. Different from 2D MOT, it is easier for trackers to predict accurate velocities in the world coordinate. We propose a complementary 3D motion prediction strategy to address the problem of object abrupt motions and short-term disappearing. More precise motion prediction tends to achieve more reliable association results and brings gain to the tracking performance. Previous works either use the detected velocities \cite{yin2021center,chen2022polar} or Kalman filter \cite{weng20203d,weng2020ab3dmot} for motion prediction. However, detected velocities have difficulty in performing long-term association as it lacks historical motion information. On the other hand, the Kalman filter produces smoother motion prediction as it leverages historical information. But when encountering abrupt and unpredictable motions or low frame rate videos, it fails to predict accurate locations. We propose a complementary motion prediction method by combining the detected object velocity with Kalman filter. Specifically, we utilize the detected velocity to perform short-term association, which is more robust to abrupt motions. We adopt the Kalman filter to predict a smoother location for each tracklet in each frame. When short-term disappearing happens, Kalman filter can maintain the object location and perform long-term association when the object reappears again. 

In summary, we propose \modelname to solve the 2D and 3D MOT problems. This builds upon our initial work ByteTrack \cite{zhang2022bytetrack}, named for each detection box is a basic unit of the tracklet, as a byte in the computer program. ByteTrack focuses on how to leverage the low-score detection boxes to reduce true object missing and fragmented trajectories in the data association strategy. The key contributions and extended results of \modelname are highlighted below. 

\myparagraph{Unified 2D and 3D Data Association.} We propose a unified data association strategy to solve the 2D and 3D MOT problems. It mines the true objects in low-score detection boxes and alleviates the problems of object missing and fragmented trajectories. Furthermore, it is non-parametric and can be combined with various detectors, making it appealing in real applications.

\myparagraph{Complementary 3D Motion Prediction.} We propose a complementary 3D motion prediction strategy to deal with the challenges of abrupt motions and short-term disappearing of objects. Specifically, we utilize the object velocity predicted by the detector to perform short-term association, which is more robust to abrupt motions. We also adopt the Kalman filter to predict a smoother location for each tracklet and perform long-term association when objects lost and reappear. 

\myparagraph{Thorough Experiments on 3D MOT Benchmarks under Different Modalities. } We conduct detailed experiments on the large-scale nuScenes dataset. The detection-driven hierarchical data association and integrated 3D motion prediction strategies are verified in 3D scenarios. \modelname achieves state-of-the-art performance on the nuScenes tracking task under both camera and LiDAR settings.

%% file: 2relatedwork.tex
In this section, we briefly review the existing works which are related to our topic including 2D object detection, 3D object detection, 2D multi-object tracking, and 3D multi-object tracking. We also discuss the relationship between these tasks.

\subsection{2D Object Detection}
2D object detection aims at predicting bounding boxes from image input. It is one of the most active topics in computer vision and it is the basis of multi-object tracking. With the rapid development of object detection \cite{ren2015faster,he2017mask,redmon2018yolov3,lin2017focal,cai2018cascade,sun2021sparse,peize2020onenet,wang2022yolov7}, more and more multi-object tracking methods begin to utilize more powerful detectors to obtain higher tracking performance. The one-stage object detector RetinaNet \cite{lin2017focal} begins to be adopted by several methods such as \cite{lu2020retinatrack,peng2020chained}. The anchor-free detector CenterNet \cite{zhou2019objects} is the most popular detector adopted by most methods \cite{zhou2020tracking,zhang2020fairmot,wu2021track,zheng2021improving,wang2020joint,tokmakov2021learning,wang2021multiple} for its simplicity and efficiency. The YOLO series detectors \cite{redmon2018yolov3,bochkovskiy2020yolov4,ge2021yolox} are also adopted by a large number of methods \cite{wang2020towards,liang2020rethinking,liang2021one,chu2021transmot,unicorn,zhang2022robust} for its excellent balance of accuracy and speed. Recently, transformer-based detectors \cite{carion2020end,zhu2020deformable,meng2021conditional} are utilized by some trackers \cite{sun2020transtrack,meinhardt2022trackformer,zeng2021motr} for their elegant end-to-end framework. We adopt YOLOX \cite{ge2021yolox} as our 2D object detector for its high efficiency. 

The MOT17 dataset \cite{milan2016mot16} provides detection results obtained by popular detectors such as DPM \cite{felzenszwalb2008discriminatively}, Faster R-CNN \cite{ren2015faster} and SDP \cite{yang2016exploit}. A large number of multi-object tracking methods \cite{xu2019spatial,chu2019famnet,bergmann2019tracking,chen2018real,zhu2018online,braso2020learning,hornakova2020lifted} focus on improving the tracking performance based on these given detection results. We also evaluate our tracking algorithm under this public detection setting. 

\subsection{3D Object Detection}
3D object detection aims at predicting three-dimensional rotated bounding boxes from images or LiDAR input. It is an indispensable component of 3D multi-object tracking as the quality of the predicted 3D bounding boxes plays an important role in the tracking performance.

LiDAR-based 3D object detection methods \cite{zhou2018voxelnet,yan2018second,lang2019pointpillars,shi2019pointrcnn,yin2021center,du2021ago,liu2022spatial} achieve impressive performance because the point clouds retrieved from LiDAR sensors contain accurate 3D structural information. However, the high cost of LiDAR limits its widespread application. 

Alternatively, recent progress in camera-based methods makes low-cost mobility widely available and thus the camera-based approaches have received increasing attention due to their low cost and rich context information. 
Due to a lack of accurate depth, the monocular 3D object detection methods \cite{chen20153d,chen2016monocular,wang2021fcos3d,zou2021devil,zhang2021objects,ye2022rope3d} directly infer geometric knowledge via deep neural networks or investigate pixel-wise depth estimation distribution to convert the image into pseudo-LiDAR points \cite{reading2021categorical,ye2020monocular, wang2019pseudo,weng2019monocular,you2019pseudo,vianney2019refinedmpl}. 

Multi-camera 3D object detection \cite{philion2020lift,li2022bevformer,wang2022detr3d,liu2022petr,li2022bevdepth,xiong2023cape} is trending and drawing extensive attention by learning powerful representations in bird's-eye-view (BEV), which is straightforward due to its unified representation and easy adaptation for downstream tasks such as future prediction and planning. Thus, the vision-centric multi-view BEV perception approaches significantly narrow the performance gap between camera-based and LiDAR-based methods. 

%Taking the pros of each modality into consideration, multi-modality fusion-based methods fully explore the data-level\cite{vora2020pointpainting,wang2021pointaugmenting} and feature-level fusion\cite{huang2020epnet,chen2022futr3d,bai2022transfusion,li2022deepfusion,liu2022bevfusion} to preserve both geometric and semantic information and thus achieve the overwhelming performance. 

LiDAR-based detectors are popular choices for 3D MOT. PointRCNN \cite{shi2019pointrcnn} and CenterPoint \cite{yin2021center} are adopted by many 3D MOT methods \cite{weng20203d,yin2021center,benbarka2021score,pang2021simpletrack,wang2021immortal} for their simplicity and effectiveness. Recently, image-based 3D object detectors \cite{kundu20183d,wang2022detr3d,li2022bevdepth} begin to be adopted by some 3D trackers such as \cite{hu2022monocular,zhang2022mutr3d,yang2022quality} because image information can provide appearance cues for tracking. Our tracking framework is modality-agnostic and thus can be easily incorporated with various 3D object detectors.

\subsection{2D Multi-Object Tracking}
Data association is the core of multi-object tracking, which first computes the similarity between tracklets and detection boxes and leverages different strategies to match them according to the similarity. 

In 2D MOT, image information serves a basic role to compute similarity. Location, motion and appearance are useful cues for data association. SORT \cite{bewley2016simple} combines location and motion cues in a very simple way. It first adopts Kalman filter \cite{kalman1960new} to predict the location of the tracklets in the new frame and then computes the IoU between the detection boxes and the predicted boxes as the similarity. Other methods \cite{zhou2020tracking,sun2020transtrack,wu2021track,shuai2021siammot} design networks to learn object motions and achieve more robust results in cases of large camera motion or low frame rate. Location and motion similarities are both accurate in the short-term association, while appearance similarity is helpful for the long-term association. An object can be re-identified using appearance similarity after being occluded for a long period of time. Appearance similarity can be measured by the cosine similarity of the Re-ID features. DeepSORT \cite{wojke2017simple} adopts a stand-alone Re-ID model to extract appearance features from the detection boxes. Recently, joint detection and Re-ID models \cite{wang2020towards,zhang2020fairmot,lu2020retinatrack,zhang2021voxeltrack,pang2021quasi,zhou2022global,xu2021segment} becomes more and more popular because of their simplicity and efficiency. 

After similarity computation, matching strategy assigns identities for the objects. This can be done by Hungarian Algorithm \cite{kuhn1955hungarian} or greedy assignment \cite{zhou2020tracking}. SORT \cite{bewley2016simple} matches the detection boxes to the tracklets by matching once. DeepSORT \cite{wojke2017simple} proposes a cascaded matching strategy that first matches the detection boxes to the most recent tracklets and then to the lost ones. MOTDT \cite{chen2018real} first utilizes appearance similarity to match and then utilizes IoU similarity to match the unmatched tracklets. QDTrack  \cite{pang2021quasi} turns the appearance similarity into probability by a bi-directional softmax operation and adopts a nearest neighbor search to accomplish matching. Attention mechanism \cite{vaswani2017attention} can directly propagate boxes between frames and perform association implicitly. Recent methods such as \cite{meinhardt2022trackformer,zeng2021motr,zhao2022tracking} propose track queries to predict the location of the tracked objects in the following frames. The matching is implicitly performed in the attention interaction process without using Hungarian Algorithm.

% Tracking can also adopted to help obtain more accurate detection results. Some methods \cite{sanchez2016online,zhu2018online,chu2019famnet,chu2019online,chu2021transmot,chen2018real} utilize single object tracking (SOT) \cite{bertinetto2016fully} or Kalman filter \cite{kalman1960new}  to predict the location of the tracklets in the following frame and fuse the predicted boxes with the detection boxes to enhance the detection results. Other methods \cite{zhang2018integrated,liang2021one} leverage tracked boxes in the previous frames to enhance feature representation of the following frame. Recently, Transformer-based \cite{vaswani2017attention,dosovitskiy2020vit,wang2021pvt,liu2021swin} detectors \cite{carion2020end,zhu2020deformable} are adopted by several methods \cite{sun2020transtrack,meinhardt2022trackformer,zeng2021motr,cai2022memot} for its strong ability to propagate boxes between frames. These methods utilize the temporal information to obtain more coherent detection results. Our method also utilize the similarity with tracklets to strength the reliability of detection boxes. 

Most 2D MOT methods focus on how to design better association strategies. However, we argue that the way detection boxes are utilized determines the upper bound of data association. After obtaining the detection boxes by various detectors, most methods \cite{wang2020towards,zhang2020fairmot,pang2021quasi,lu2020retinatrack} only keep the high score boxes by a threshold, \ie 0.5, and use those boxes as the input of data association. This is because the low score boxes contain many backgrounds which harm the tracking performance. However, we observe that many occluded objects can be correctly detected but have low scores. To reduce missing detections and keep the persistence of trajectories, we keep all the detection boxes and associate across every one of them. We focus on how to make full use of detection boxes from high scores to low ones in the association process. 

\subsection{3D Multi-Object Tracking}
3D MOT shares many commonalities with 2D MOT, \ie data association. Most 3D MOT methods follow the tracking-by-detection paradigm, which first detects objects and then associates them across time. Compared with 2D MOT, the location and motion cues utilized in 3D MOT are more accurate and reliable because they contain depth information. For example, when two pedestrians come across each other, the 2D IoU obtained from the image plane is large and it is difficult to distinguish them via the 2D location. In 3D scenarios, the two pedestrians can be easily separated according to their differences in depth. 

LiDAR-based 3D trackers tend to utilize location and motion cues to compute similarity. Similar to SORT \cite{bewley2016simple}, AB3DMOT \cite{weng20203d} provides a simple baseline and a new evaluation metric for 3D MOT, which adopts Kalman filter as motion model and use 3D IoU between detections and tracklets for association. CenterPoint \cite{yin2021center} extends the center-based tracking paradigm in CenterTrack \cite{zhou2020tracking} to 3D. It utilizes the predicted object velocity as the constant velocity motion model and shows effectiveness under abrupt motions. Following works \cite{benbarka2021score,chiu2021pro,pang2021simpletrack,wang2021immortal} focus on the improvement of association metrics and life cycle management. QTrack \cite{yang2022quality} estimates the quality of predicted object attributes and proposes a quality-aware association strategy for more robust association. 

Visual appearance features extracted from images can further enhance the long-term association in 3D MOT. QD3DT \cite{hu2022monocular} proposes a quasi-dense similarity learning of appearance features following QDTrack \cite{pang2021quasi} in 2D MOT to handle the object reappearance problems. GNN3DMOT \cite{weng2020gnn3dmot} integrates motion and appearance features and learns the interaction between them via graph neural networks. TripletTrack \cite{marinello2022triplettrack} proposes local object feature embeddings to encode information about the visual appearance and monocular 3D object characteristics to achieve more robust performance in case of occlusions and missing detections. Recent progress in Transformer \cite{vaswani2017attention} makes it appealing in 3D MOT. Following MOTR \cite{zeng2021motr} in 2D MOT, camera-based 3D tracker MUTR3D \cite{zhang2022mutr3d} utilizes Transformer to learn 3D representations with 2D visual information and propagate the 3D bounding boxes across time in an end-to-end manner. 

We only utilize the motion cues to perform data association for a simpler unification of 2D and 3D MOT. We propose an integrated motion prediction strategy based on the detected velocity and Kalman filter. The smooth location predicted by Kalman filter is adopted for long-term association, which plays a similar role to deal with the reappearance problems as appearance features.

%% file: 3bytetrack.tex
\begin{figure*}
\centering
\includegraphics[width=0.95\textwidth]{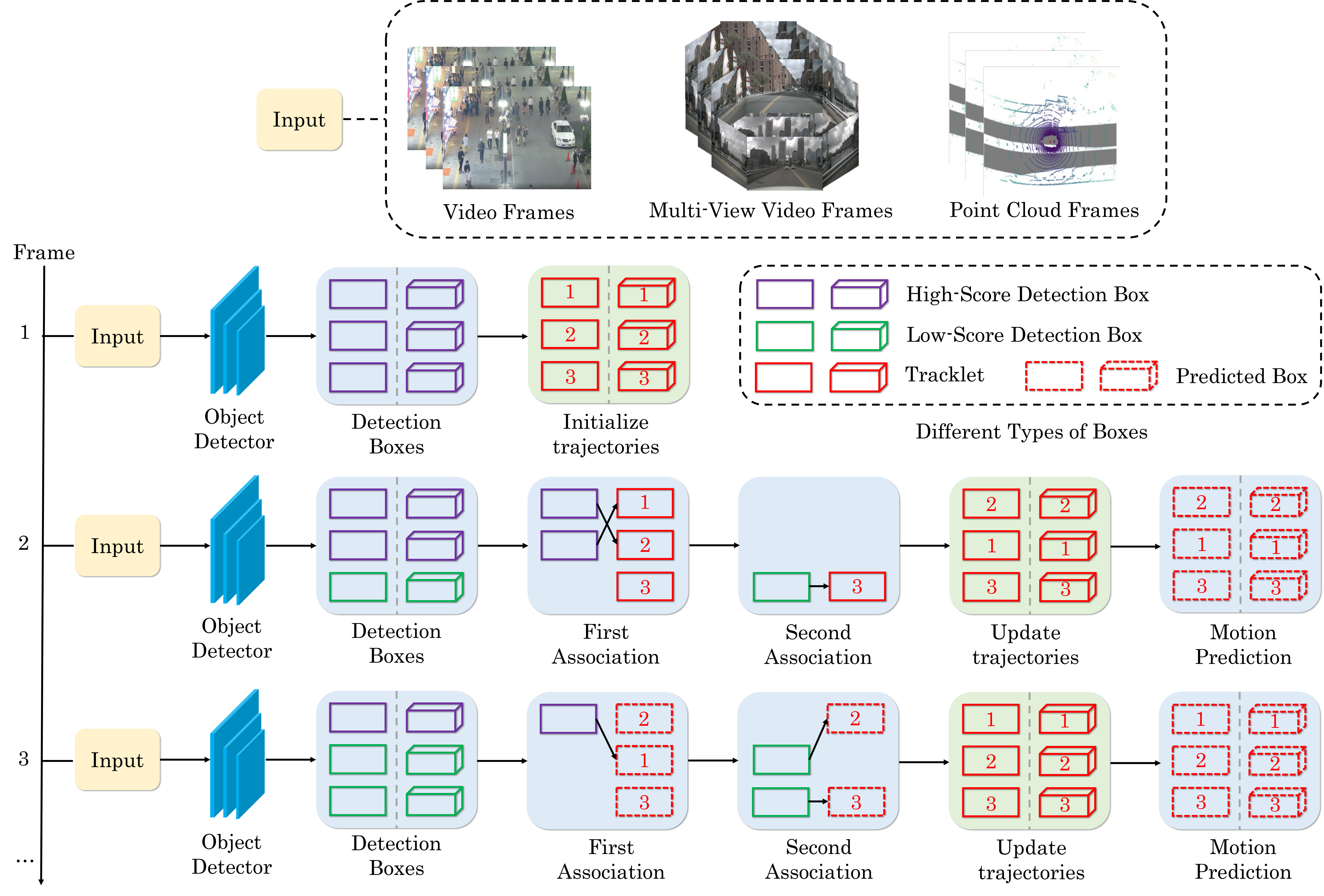}
\caption{Overview of our unified 2D and 3D MOT framework \modelname. The input can be images, multi-view images, or point clouds. We first adopt a 2D or 3D object detector to obtain the detection boxes. In the first frame, we initialize all the high-score detection boxes as tracklets. In the following frames, we first associate the high-score detection boxes with all the tracklets and then associate the low-score detection boxes with the unmatched tracklets. The associations are performed between the detection boxes and the tracklets after motion prediction. The final outputs are the updated trajectories in each frame. }
\label{fig:pipeline}
\vspace{-3mm}
\end{figure*}

We solve the problems of 2D and 3D MOT in a simple and unified framework. It contains three parts: object detection, motion prediction, and data association. We first introduce how we phrase the problems of 2D MOT and 3D MOT in Sec.~\ref{subsec:problem}. Then we list some 2D and 3D detectors along with the basic Kalman filter motion model in Sec.~\ref{subsec:preliminary}. In Sec.~\ref{subsec:motion}, we introduce the complementary 3D motion prediction strategy proposed specially for 3D MOT. Finally, we elaborate on the core steps of the proposed detection-driven hierarchical data association strategy and how it alleviates the problem of object missing and fragmented trajectories in Sec.~\ref{subsec:asso}. The overview of the tracking framework is shown in Fig.~\ref{fig:pipeline}.

\subsection{Problem Formulation}
\label{subsec:problem}
\myparagraph{Multi-object tracking. } The goal of multi-object tracking is to estimate object trajectories in videos. Suppose we are going to obtain $L$ trajectories $\mathbb{S} = \{s^1,s^2,...,s^L\}$ in a video. Each trajectory $s^i=\{\mathbf{b}^i_{t_1},\mathbf{b}^i_{t_1+1},...,\mathbf{b}^i_{t_2}\}$ contains the location information of an object within a time period, \ie from frame $t_1$ to frame $t_2$, where the object appears. In 2D MOT, the location of object $i$ at frame $t$ can be represented as ${\mathbf{b}^i_{t}}_{2d}=[x_1,y_1,x_2,y_2] \in \mathbb{R}^4$, where $(x_1,y_1)$, $(x_2,y_2)$ are the top-left and bottom-right coordinates of the 2D object bounding box in the image plane. In 3D MOT, the tracking procedure is usually performed in the 3D world coordinates. The 3D location of object $i$ at frame $t$ can be denoted as ${\mathbf{b}^i_{t}}_{3d}=[x,y,z,\theta,l,w,h] \in \mathbb{R}^7$, where $(x,y,z)$ are the 3D world locations of the object center, $\theta$ is the object orientation and $(l,w,h)$ are the object dimensions.

\myparagraph{Data association. } We follow the popular \textit{tracking-by-detection} paradigm in multi-object tracking, which first detects objects in individual video frames and then associates them between frames and forms trajectories over time. Suppose we have $M$ detections and $N$ history trajectories at frame $t$, our goal is to assign each detection to one of the trajectories, which has the same identity in an entire video. Let $\mathbb{A}$ denote the space which consists of all possible associations (or matchings). Under the setting of multi-object tracking, each detection matches at most one trajectory and each trajectory matches at most one detection. We define the space $\mathbb{A}$ as follow:
\begin{align}
\label{eq:space}
    \mathbb{A} &= \Big \{A = \Big (m_{ij}\Big )_{i\in \mathbb{M},j\in \mathbb{N}} \; \Big | \; m_{ij} \in \{0,1\} \\
               & \land \; \sum_{i=0}^{M} m_{ij} \leqslant 1, \forall j \in \mathbb{N} \\
               & \land \; \sum_{j=0}^{N} m_{ij} \leqslant 1, \forall i \in \mathbb{M} \Big \},
\end{align}
where $\mathbb{M}=\{1,2,...,M\}$, $\mathbb{N}=\{1,2,...,N\}$ and $A$ is one possible matching of the entire $M$ detections and $N$ trajectories. When the $i_{th}$ detection is matched to the $j_{th}$ trajectory, then $m_{ij}=1$. Let $\mathbf{d}^1_t,...,\mathbf{d}^M_t$ and $\mathbf{h}^1_t,...,\mathbf{h}^N_t$ be the locations of all $M$ detections and $N$ trajectories at frame $t$, respectively. We compute a similarity matrix $\mathbf{S}_t \in \mathbb{R}^{M\times N}$ between all the detections and trajectories as follows:
\begin{equation}
\label{eq:sim}
    \mathbf{S}_t(i,j) = \operatorname{sim}(\mathbf{d}_t^i,\mathbf{h}_t^j),
\end{equation}
where the similarity can be computed by some spatial distances between the detections and the trajectories such as IoU or L2 distance. Our goal is to obtain the optimal matching $A^*$ where the total similarities (or scores) between the matched detections and trajectories is the highest:
\begin{equation}
\label{eq:optimal}
    A^* = \mathop{\arg\max}\limits_{m_{ij}=1}\sum_{i\in\mathbb{M},j\in\mathbb{N}}\mathbf{S}_t(i,j),
\end{equation}

\subsection{Preliminary}
\label{subsec:preliminary}
\myparagraph{2D object detector. } We adopt YOLOX \cite{ge2021yolox} as our 2D object detector. YOLOX is an anchor-free detector equipped with advanced detection techniques, \ie, a decoupled head, and the leading label assignment strategy SimOTA derived from OTA \cite{ge2021ota}. It also adopts strong data augmentations such as mosaic \cite{bochkovskiy2020yolov4} and mix-up \cite{zhang2017mixup} to further enhance the detection performance. YOLOX achieves an excellent balance of speed and accuracy compared with other modern detectors \cite{zhang2022dino,liu2021swin} and is appealing in real applications. 

\myparagraph{Camera-based 3D object detector. } We follow the multi-camera 3D object detection setting, which shows advantages over monocular methods by learning powerful and unified representations in bird's-eye-view (BEV). We utilize PETRv2 \cite{liu2022petrv2} as our camera-based 3D object detector. It is built upon PETR \cite{liu2022petr}, which extends the transformer-based 2D object detector DETR \cite{carion2020end} to multi-view 3D setting by encoding the position information of 3D coordinates into image features. PETRv2 utilizes the temporal information of previous frames to boost detection performance. 

\myparagraph{LiDAR-based 3D object detector. } We adopt CenterPoint \cite{yin2021center} and TransFusion-L \cite{bai2022transfusion} as our LiDAR-based 3D object detectors. CenterPoint utilizes a keypoint detector to find the centers of objects, and simply regresses to other 3D attributes. It also refines these 3D attributes using additional point features on the object in a second stage. TransFusion-L consists of convolutional backbones and a detection head based on a transformer decoder. It predicts 3D bounding boxes from a LiDAR point cloud using a sparse set of object queries. 

\myparagraph{Basic motion model. } We utilize Kalman filter \cite{kalman1960new} as our basic motion model for both 2D and 3D MOT. Similar to \cite{wojke2017simple}, we define an eight-dimensional state space $(u,v,a,b,\dot{u},\dot{v},\dot{a},\dot{b})$ in 2D tracking scenario, where $\mathbf{P}^{2d}=(u,v,a,b)$ are the 2D bounding box center position, the aspect ratio (width / height) and the bounding box height. $\mathbf{V}^{2d}=(\dot{u},\dot{v},\dot{a},\dot{b})$ are the respective velocities in the image plane. In 3D tracking scenario, we follow \cite{weng20203d} to define a ten-dimensional state space $(x,y,z,\theta,l,w,h,\dot{x},\dot{y},\dot{z})$, where $\mathbf{P}^{3d}=(x,y,z)$ is the 3D bounding box center position, $(l,w,h)$ is the object's size, $\theta$ is the object orientation and $\mathbf{V}^{3d}=(\dot{x},\dot{y},\dot{z})$ are the respective velocities in the 3D space. Different from \cite{weng20203d}, we define the state space in the 3D world coordinates to eliminate the effects of ego-motion. We directly adopt a standard Kalman filter with constant velocity motion and linear observation model. The motion prediction process at frame $t+1$ in 2D and 3D tracking scenario can be denoted as follows:
% \begin{equation}
% \begin{split}
%     \label{eq:predict}
%     u_{t+1}=u_{t}+\dot{u}_{t} \qquad v_{t+1}=v_{t}+\dot{v}_{t} \\
%     a_{t+1}=a_{t}+\dot{a}_{t} \qquad b_{t+1}=b_{t}+\dot{b}_{t},
% \end{split}
% \end{equation}
% It is similar in 3D tracking scenario:
% \begin{equation}
% \begin{split}
%     \label{eq:predict}
%     x_{t+1}=x_{t}+\dot{x}_{t} \quad
%     y_{t+1}=y_{t}+\dot{y}_{t} \quad
%     z_{t+1}=z_{t}+\dot{z}_{t}.
% \end{split}
% \end{equation}
\begin{equation}
\begin{split}
    \label{eq:predict}
    \mathbf{P}^{2d}_{t+1} = \mathbf{P}^{2d}_{t} + \mathbf{V}^{2d}_{t} \\
    \mathbf{P}^{3d}_{t+1} = \mathbf{P}^{3d}_{t} + \mathbf{V}^{3d}_{t}
\end{split}
\end{equation}
The updated state of each trajectory is the weighted average of the trajectory and the matched detection (or observation). The weights are determined by the uncertainty of both the trajectory and the matched detection following the Bayes rule.

\subsection{Complementary 3D Motion Prediction}
\label{subsec:motion}
We propose a complementary 3D motion prediction strategy to address the abrupt motions and short-term object disappearing problems in the driving scenarios. Specifically, we adopt the detected velocity for short-term association and the Kalman filter for long-term association.

%For example, the velocities of vehicles change drastically when cornering and braking occur. Linear motion models such as Kalman filter \cite{kalman1960new} have difficulty handling these situations. Another case is that pedestrians tend to get lost for a period of time when occlusion happens. Re-associating these targets needs long-term motion prediction utilizing history information.  

\begin{figure}[!h]
	\centering
	\includegraphics[width=0.98\linewidth]{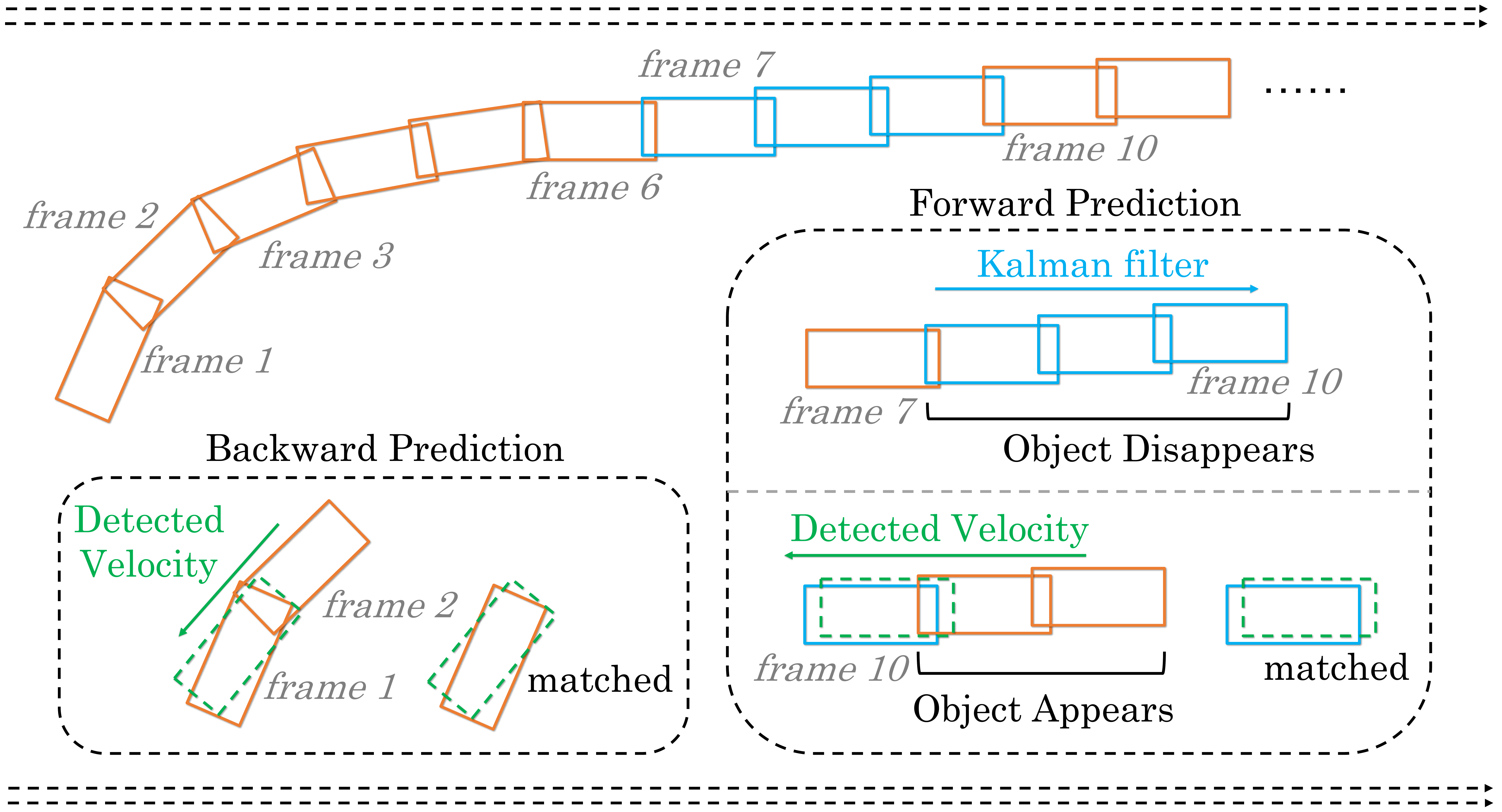}
	\caption{Illustration of the complementary motion prediction strategy. Boxes in orange represent the detection results. Boxes in blue represent the predicted location by the Kalman filter. Dashed boxes in green represent the location of backward prediction by the detected velocity. }
	\label{fig:motion}
	\vspace{-5mm}
\end{figure}

In 3D scenarios, modern detectors \cite{li2022bevformer,liu2022petrv2,yin2021center} have the ability to predict accurate short-term velocity via temporal modeling. Kalman filter models the smooth long-term velocity via state updating based on historical information. We maximize the strengths of both motion models through a bilateral prediction strategy. We adopt the Kalman filter to perform forward prediction and adopt the detected object velocity to perform backward prediction. The backward prediction is responsible for the short-term association of alive tracklets while the forward prediction is leveraged for the long-term association of lost tracklets. Fig.~\ref{fig:motion} illustrates the complementary motion prediction strategy. 

Suppose we have $M$ detections $\mathbf{D}^t \in \mathbb{R}^{M \times 7}$ and the detected object velocity $\mathbf{V}^t \in \mathbb{R}^{M \times 2}$ in both $x$ and $y$ directions at frame $t$. The 
backward prediction can be computed as follows:
\begin{equation}
\label{eq:back}
\hat{\mathbf{D}}^{t-1}_x=\mathbf{D}^t_x - \mathbf{V}^t_x \qquad
\hat{\mathbf{D}}^{t-1}_y=\mathbf{D}^t_y - \mathbf{V}^t_y,
\end{equation}
Suppose there are $N$ tracklets $\mathbf{T}^{t-1} \in \mathbb{R}^{N \times 7}$ at frame $t-1$, We adopt the Kalman filter described in Sec.~\ref{subsec:preliminary} to perform forward prediction as follows:
\begin{equation}
\begin{split}
    \label{eq:forward}
    \mathbf{T}^{t}_{x,y,z}=\mathbf{T}^{t-1}_{x,y,z} + \dot{\mathbf{T}}^{t-1}_{x,y,z},
\end{split}
\end{equation}
where $\dot{\mathbf{T}}^{t-1}_{x,y,z}$ is the smoothed velocity in $x$, $y$ and $z$ directions computed by the Kalman filter in Sec.~\ref{subsec:preliminary}. 

We utilize the unified 2D and 3D data association strategy introduced in Sec.~\ref{subsec:asso} after the bilateral prediction. In the first association, the similarity $\mathbf{S}_t \in \mathbb{R}^{M\times N}$ is computed between the detection results $\mathbf{D}^{t-1}$ from backward prediction and the tracklets $\mathbf{T}^{t-1}$ as follows:
\begin{equation}
\label{eq:sim3d}
    \mathbf{S}_{t}(i,j) = \operatorname{GIoU}(\mathbf{D}_i^{t-1},\mathbf{T}_j^{t-1}),
\end{equation}
We adopt 3D GIoU \cite{rezatofighi2019generalized} as the similarity metric to address the occasional non-overlapping problem between the detection boxes and the tracklet boxes. We use Hungarian algorithm \cite{kuhn1955hungarian} to complete the identity assignment based on $\mathbf{S}_t$.
After the association, the matched detections $\mathbf{D}^t_{match}$ are utilized to update the matched tracklets $\mathbf{T}^t_{match}$ following the standard Kalman filter updating rule. The forward prediction strategy in Eq.~\ref{eq:forward} plays an important role when a tracklet gets lost, \ie with no matched detections. When the lost object reappears in the following frames, it can be re-associated via the similarity with the predicted location, which is also called track rebirth. The second association in Algorithm~\ref{algo:byte} follows the same procedure as the first association. 

We adopt the detection scores to further enhance motion prediction by adaptively updating the measurement uncertainty matrix $\hat{\mathbf{R}}_t^j$ in the Kalman filter of trajectory $j$ at frame $t$ as follows:
\begin{equation}
\label{eq:update}
    \hat{\mathbf{R}}_t^j = \alpha (1-s^j_t)^2 \mathbf{R}_t^j ,
\end{equation}
where $s^j_t$ is the detection score of trajectory $j$ at frame $t$ and $\alpha$ is a hyperparameter that controls the magnitude of the uncertainty. By plugging the detection score into the uncertainty matrix, we make the Kalman filter more robust to detections of different qualities. 

\subsection{Unified 2D and 3D Data Association}
\label{subsec:asso}
We propose a simple, effective, and unified data association method for 2D and 3D MOT. Different from previous methods \cite{wang2020towards,zhang2020fairmot,pang2021quasi,weng20203d,yin2021center}, which only keep the high-score detection boxes, we keep every detection box and separate them into high-score ones and low-score ones. The whole pipeline of our detection-driven hierarchical data association strategy is shown in Figure~\ref{fig:pipeline}. 

\myparagraph{Overview. }In the first frame of the video, we initialize all the detection boxes as tracklets. In the following frames, we first associate the high-score detection boxes with the tracklets. Some tracklets get unmatched because they do not match an appropriate high-score detection box, which usually happens when occlusion, motion blur, or size change occurs. We then associate the low-score detection boxes and these unmatched tracklets to recover the objects in low-score detection boxes and filter out the background, simultaneously. The pseudo-code of \modelname is shown in Algorithm~\ref{algo:byte}.

\begin{algorithm}[!h]
\SetAlgoLined
\DontPrintSemicolon
\SetNoFillComment
\footnotesize
\KwIn{A video sequence $\texttt{V}$; object detector $\texttt{Det}$; detection score threshold {$\tau$}}
\KwOut{Tracks $\mathcal{T}$ of the video}

Initialization: $\mathcal{T} \leftarrow \emptyset$\;
\For{frame $f_k$ in $\texttt{V}$}{
	\tcc{Figure 3(a)}
	\tcc{predict detection boxes \& scores}
	$\mathcal{D}_k \leftarrow \texttt{Det}(f_k)$ \;
	$\mathcal{D}_{high} \leftarrow \emptyset$ \;
	\textcolor{codegreen}{$\mathcal{D}_{low} \leftarrow \emptyset$} \;
	\For{$d$ in $\mathcal{D}_k$}{
	\If{$d.score > \tau$}{
	$\mathcal{D}_{high} \leftarrow  \mathcal{D}_{high} \cup \{d\}$ \;
	}
	\Else{
	\textcolor{codegreen}{$\mathcal{D}_{low} \leftarrow  \mathcal{D}_{low} \cup \{d\}$ \;
	}}
	}
	
    \BlankLine	
	\BlankLine
	\tcc{predict new locations of tracks}
	\For{$t$ in $\mathcal{T}$}{
	$t \leftarrow \texttt{MotionPredictor}(t)$ \;
	}
	
    \BlankLine
    \BlankLine
	% \tcc{Figure 3(b)}
	\tcc{first association}
	Associate $\mathcal{T}$ and $\mathcal{D}_{high}$ using \texttt{Similarity\#1}\;
	$\mathcal{D}_{remain} \leftarrow \text{remaining object boxes from } \mathcal{D}_{high}$ \;
	$\mathcal{T}_{remain} \leftarrow \text{remaining tracks from } \mathcal{T}$ \;
	
	\BlankLine
	\BlankLine
	% \tcc{Figure 3(c)}
    \tcc{second association}
	\textcolor{codegreen}{
	Associate $\mathcal{T}_{remain}$ and $\mathcal{D}_{low}$ using \texttt{Similarity\#2}\;}
	\textcolor{codegreen}{
	$\mathcal{T}_{re-remain} \leftarrow \text{remaining tracks from } \mathcal{T}_{remain}$ \;}

    \BlankLine
	\BlankLine
	\tcc{delete unmatched tracks}
	$\mathcal{T} \leftarrow \mathcal{T} \setminus \mathcal{T}_{re-remain}$ \;
	
    \BlankLine
	\BlankLine
	\tcc{initialize new tracks}
    \For{$d$ in $\mathcal{D}_{remain}$}{
	$\mathcal{T} \leftarrow  \mathcal{T} \cup \{d\}$ \;
	}
}
Return: $\mathcal{T}$
\caption{Pseudo-code of \modelname.}
\algorithmfootnote{Track rebirth~\cite{wojke2017simple,zhou2020tracking} is not shown in the algorithm for simplicity. In \textcolor{codegreen}{green} is the key of our method. }
\label{algo:byte}
\end{algorithm}

% \vspace{-5mm}

\myparagraph{Input. }The input of \modelname is a video sequence $\texttt{V}$, along with an object detector $\texttt{Det}$. We also set a detection score threshold $\tau$. The output is the tracks $\mathcal{T}$ of the video and each track contains the bounding box and identity of the object in each frame.

\myparagraph{Detection boxes. }For each frame in the video, we predict the detection boxes and scores using the detector $\texttt{Det}$. We separate all the detection boxes into two parts $\mathcal{D}_{high}$ and $\mathcal{D}_{low}$ according to the detection score threshold $\tau$. For the detection boxes whose scores are higher than $\tau$, we put them into the high-score detection boxes $\mathcal{D}_{high}$. For the detection boxes whose scores are lower than $\tau$, we put them into the low-score detection boxes $\mathcal{D}_{low}$ (Line 3 to 13 in Algorithm~\ref{algo:byte}). 

\myparagraph{Motion prediction. }After separating the low-score detection boxes and the high-score detection boxes, we predict the new locations in the current frame of each track in $\mathcal{T}$ (Line 14 to 16 in Algorithm~\ref{algo:byte}) For 2D MOT, we directly adopt the Kalman filter for motion prediction. For 3D MOT, we utilize the complementary motion prediction strategy introduced in Sec.~\ref{subsec:motion}.

\myparagraph{High-score boxes association. }The first association is performed between the high-score detection boxes $\mathcal{D}_{high}$ and all the tracks $\mathcal{T}$ (including the lost tracks $\mathcal{T}_{lost}$). \texttt{Similarity\#1} can be computed by the spatial distance such as IoU between the detection boxes $\mathcal{D}_{high}$ and the predicted box of tracks $\mathcal{T}$. Then, we adopt Hungarian Algorithm \cite{kuhn1955hungarian} to finish the matching based on the similarity. We keep the unmatched detections in $\mathcal{D}_{remain}$ and the unmatched tracks in $\mathcal{T}_{remain}$ (Line 17 to 19 in Algorithm~\ref{algo:byte}). 

The whole pipeline is highly flexible and can be compatible to other different association methods. For example, when it is combined with FairMOT~\cite{zhang2020fairmot}, Re-ID feature is added into \texttt{* first association *} in Algorithm~\ref{algo:byte}, others are the same. In the experiments of 2D MOT, we apply the association method to 9 different state-of-the-art trackers and achieve notable improvements on almost all the metrics.

\myparagraph{Low-score boxes association. }The second association is performed between the low-score detection boxes $\mathcal{D}_{low}$ and the remaining tracks $\mathcal{T}_{remain}$ after the first association. We keep the unmatched tracks in $\mathcal{T}_{re-remain}$ and just delete all the unmatched low score detection boxes since we view them as background. (line 20 to 21 in Algorithm~\ref{algo:byte}). We find it important to use IoU alone as the \texttt{Similarity\#2} in the second association because the low-score detection boxes usually contain severe occlusion or motion blur and the appearance features are not reliable. Thus, when applied to other Re-ID-based trackers \cite{wang2020towards,zhang2020fairmot,pang2021quasi}, we do not adopt appearance similarity in the second association. 

\myparagraph{Track rebirth. }After the association, the unmatched tracks will be deleted from the tracklets. We do not list the procedure of track rebirth \cite{wojke2017simple,chen2018real,zhou2020tracking} in Algorithm~\ref{algo:byte} for simplicity. Actually, it is necessary for the long-term association to preserve the identity of the tracks. For the unmatched tracks $\mathcal{T}_{re-remain}$ after the second association, we put them into $\mathcal{T}_{lost}$. For each track in $\mathcal{T}_{lost}$, only when it exists for more than a certain number of frames, \ie 30, we delete it from the tracks $\mathcal{T}$. Otherwise, we remain the lost tracks $\mathcal{T}_{lost}$ in $\mathcal{T}$ (Line 22 in Algorithm~\ref{algo:byte}). Finally, we initialize new tracks from the unmatched high-score detection boxes $\mathcal{D}_{remain}$ after the first association (Line 23 to 27 in Algorithm~\ref{algo:byte}). The output of each individual frame is the bounding boxes and identities of the tracks $\mathcal{T}$ in the current frame.

\myparagraph{Discussion. }We empirically find that the detection score drops when the occlusion ratio increases. When an occlusion case happens, the score first decreases and then increases, as the pedestrian is occluded first and then re-appears. This inspires us to first associate the high-score boxes with the tracklets. If a tracklet does not match any of the high score boxes, it is highly possible to be occluded and the detection score drops. Then, we associate it with the low-score boxes to track the occluded target. For those false positives in the low-score boxes, no tracklets match them. Thus, we throw them away. This is the key point that our data association algorithm works.

%% file: 5datasets.tex
\subsection{Datasets}
\myparagraph{MOT17 dataset \cite{milan2016mot16}} has videos captured by both moving and stationary cameras from various viewpoints at different frame rates. It contains 7 training videos and 7 test videos. We use the first half of each video in the training set of MOT17 for training and the last half for validation following \cite{zhou2020tracking}. MOT17 provides both ``public detection'' and ``private detection'' protocols. Public detectors include DPM \cite{felzenszwalb2008discriminatively}, Faster R-CNN \cite{ren2015faster} and SDP \cite{yang2016exploit}. For the private detection setting, we train on the combination of CrowdHuman dataset \cite{shao2018crowdhuman} and MOT17 half training set following \cite{zhou2020tracking,sun2020transtrack,zeng2021motr,wu2021track} in ablation studies. We add Cityperson \cite{zhang2017citypersons} and ETHZ \cite{ess2008mobile} for training following \cite{wang2020towards,zhang2020fairmot,liang2020rethinking} when testing on the test set of MOT17. 

\myparagraph{MOT20 dataset \cite{dendorfer2020mot20}} capture the videos in very crowded scenes so there is a lot of occlusion happening. The average pedestrian in a frame is much larger than that of MOT17 (139 \textit{vs.} 33). MOT20 contains 4 training videos and 4 test videos with longer video length. It also provides public detection results from Faster-RCNN. We only use CrowdHuman dataset and the training set of MOT20 for training under the private detection setting. 

\myparagraph{HiEve dataset \cite{lin2020human}} is a large-scale human-centric dataset focusing on crowded and complex events. It contains longer average trajectory length, bringing more difficulty for human tracking tasks. HiEve captures videos on 30+ different scenarios including subway station, street, and dining hall, making the tracking problem a more challenging task. It contains 19 training videos and 13 test videos. We combine CrowdHuman and the training set of HiEve for training. 

\myparagraph{BDD100K dataset \cite{yu2020bdd100k}} is the largest 2D driving video dataset and the dataset splits of the 2D MOT task are 1400 videos for training, 200 videos for validation and 400 videos for testing. It needs to track objects of 8 classes and contains cases of large camera motion. We combine the training sets of the detection task and the 2D MOT task for training. 

\myparagraph{nuScenes dataset \cite{caesar2020nuscenes}} is a large scale 3D object detection and tracking dataset. It is the first dataset to carry the full autonomous vehicle sensor suite: 6 cameras, 5 radars and 1 lidar, all with full 360 degree field of view. The tracking task contains 3D annotations with 7 object classes. nuScenes comprises 1000 scenes, including 700 training videos, 150 validation videos and 150 test videos. For each sequence, only the key frames sampled at 2 FPS are annotated. There are about 40 key frames per camera in a sequence. We only use the key frames in each sequence for tracking. 

\subsection{Metrics}
\myparagraph{2D MOT. } We use the CLEAR metrics \cite{bernardin2008evaluating}, including MOTA, FP, FN, IDs, \textit{etc.}, IDF1 \cite{ristani2016performance} and HOTA \cite{luiten2021hota} to evaluate different aspects of the tracking performance. MOTA is computed based on FP, FN and IDs as follows:
\begin{equation}
    MOTA=1-\frac{IDS+FP+FN}{GT},
\end{equation}
where GT stands for the number of ground-truth objects. Considering the amount of FP and FN are larger than IDs, MOTA focuses more on the detection performance. IDF1 evaluates the identity preservation ability and focus more on the association performance. HOTA is a very recently proposed metric which explicitly balances the effect of performing accurate detection, association and localization. For BDD100K dataset, there are some multi-class metrics such as mMOTA and mIDF1. mMOTA / mIDF1 is computed by averaging the MOTA / IDF1 of all the classes. 

\myparagraph{3D MOT. } The nuScenes tracking benchmark adopts Average Multi-Object Tracking Accuracy (AMOTA) \cite{weng20203d} as the main metric, which averages the MOTA metric over different recall thresholds to reduce the effect of detection confidence thresholds. sMOTA$_r$ \cite{weng20203d} augments MOTA by a term to adjust for the respective recall and guarantees that the sMOTA$_r$ values range from 0.0 to 1.0:
\begin{align}
\begin{split}
    & sMOTA_r=\\
    &\operatorname{max}(0,1-\frac{IDS_r+FP_r+FN_r-(1-r)P}{rP}),
\end{split}
\end{align}
Then, 40-point interpolation is utilized to compute the AMOTA metric:
\begin{equation}
    AMOTA=\frac{1}{|\mathcal{R}|}\sum_{r \in \mathcal{R}}sMOTA_r
\end{equation}
However, we identify in experiments that it still needs to select a suitable detection score threshold as the false positives may mislead the association results.

%% file: 6experiments.tex
\subsection{Implementation Details}
\label{subsec:detail}
\myparagraph{2D MOT. } The detector is YOLOX \cite{ge2021yolox} with YOLOX-X as the backbone and COCO-pretrained model \cite{lin2014microsoft} as the initialized weights. For MOT17, the training schedule is 80 epochs on the combination of MOT17, CrowdHuman, Cityperson and ETHZ. For MOT20 and HiEve, we only add CrowdHuman as additional training data. For BDD100K, we do not use additional training data and only train 50 epochs. The input image size is 1440 $\times$ 800 and the shortest side ranges from 576 to 1024 during multi-scale training. The data augmentation includes Mosaic \cite{bochkovskiy2020yolov4} and Mixup \cite{zhang2017mixup}. The model is trained on 8 NVIDIA Tesla V100 GPU with batch size of 48. The optimizer is SGD with weight decay of $5\times10^{-4}$ and momentum of 0.9. The initial learning rate is $10^{-3}$ with 1 epoch warm-up and cosine annealing schedule. The total training time is about 12 hours. Following \cite{ge2021yolox}, FPS is measured with FP16-precision \cite{micikevicius2017mixed} and batch size of 1 on a single GPU.

For the tracking part, the default detection score threshold $\tau$ is 0.6, unless otherwise specified. For the benchmark evaluation of MOT17, MOT20 and HiEve, we only use IoU as the similarity metrics. In the linear assignment step, if the IoU between the detection box and the tracklet box is smaller than 0.2, the matching will be rejected. For the lost tracklets, we keep it for 30 frames in case it appears again. For BDD100K, we use UniTrack \cite{wang2021different} as the Re-ID model. In ablation study, we use FastReID \cite{he2020fastreid} to extract Re-ID features for MOT17. 

\myparagraph{3D MOT. } For the camera modality setting, we adopt PETRv2 \cite{liu2022petrv2} with VoVNetV2 backbone \cite{lee2020centermask} as the detector. The input image size is 1600 $\times$ 640. The detection query number is set as 1500. It is trained with the pretrained model FCOS3D \cite{wang2021fcos3d} on validation dataset and with DD3D \cite{park2021pseudo} pretrained model on the test dataset as initialization. The optimizer is AdamW \cite{loshchilov2017decoupled} with a weight decay of 0.01. The learning rate is initialized with $2.0 \times 10^{-4}$ and decayed with cosine annealing policy. The model is trained on 8 Tesla A100 GPUs with a batch size of 8 for 24 epochs. For the LiDAR modality setting, we utilize TransFusion-L with VoxelNet \cite{zhou2018voxelnet} 3D backbone as the detector for the nuScenes test set. The model is trained for 20 epochs with the LiDAR point clouds as input. For the results on the nuScenes validation set, we adopt CenterPoint \cite{yin2021center} with VoxelNet backbone as the detector. The same detection results are used for fair comparisons with other 3D MOT methods \cite{yin2021center,weng20203d,chiu2021pro,benbarka2021score}. 

For the tracking part, the detection score threshold $\tau$ is 0.2 for PETRv2 and CenterPoint. The detection score of TransFusion-L is lower than other detectors because it is computed as the geometric average of the heatmap score and the classification score, so we set the threshold $\tau$ as 0.01 for TransFusion-L. We set different 3D GIoU thresholds for different object classes as they have different sizes and velocities. Concretely, we set -0.7 for bicycle, -0.2 for bus, -0.1 for car, -0.5 for motorcycle, -0.7 for pedestrian, -0.4 for trailer and -0.1 for truck. Similar to 2D MOT, we keep the lost tracklets for 30 frames in case it appears again. The hyperparameter $\alpha$ for updating the measurement uncertainty matrix in Eq.~\ref{eq:update} is 100 for camera-based method and 10 for LiDAR-based method. 

\subsection{2D MOT}
We evaluate ByteTrack under the 2D MOT setting in this section. We first present some experimental results of ablation analysis on MOT17 datasets. Then, we compare ByteTrack with other trackers on 4 different benchmarks. We denote the hierarchical data association strategy of ByteTrack as BYTE in the following paragraphs for simplicity. 

\subsubsection{Ablation Studies}
\myparagraph{Similarity analysis. }
We choose different types of similarity for the first association and the second association of BYTE. The results are shown in Table~\ref{table_sim}. We can see that either IoU or Re-ID can be a good choice for \texttt{Similarity\#1} on MOT17. IoU achieves better MOTA and IDs while Re-ID achieves higher IDF1. It is important to utilize IoU as \texttt{Similarity\#2} in the second association because the low score detection boxes usually contains severe occlusion and thus Re-ID features are not reliable. From Table~\ref{table_sim} we can find that using IoU as \texttt{Similarity\#2} increases 0.8 MOTA compared to Re-ID, which indicates that Re-ID features of the low score detection boxes are not reliable. We choose to use IoU as similarity in both associations. It is also more unified with the 3D tracking framework.

\begin{table}[!h]
\caption{Comparison of different type of similarity metrics used in the first association and the second association on the MOT17 validation set. The best results are shown in \textbf{bold}.}
\begin{center}
{\input{tables/sim}}
\end{center}
\label{table_sim}
\vspace{-5mm}
\end{table}

\myparagraph{Comparisons with other association methods.}
We compare BYTE with other popular association methods including SORT \cite{bewley2016simple}, DeepSORT \cite{wojke2017simple} and MOTDT \cite{chen2018real} on the validation set of MOT17. Results are shown in Table~\ref{table_ass}. 

SORT can be seen as our baseline method because both methods only adopt Kalman filter to predict the object motion.  We can find that BYTE improves the MOTA metric of SORT from 74.6 to 76.6, IDF1 from 76.9 to 79.3 and decreases IDs from 291 to 159. This highlights the importance of the low score detection boxes and proves the ability of BYTE to recover object boxes from low score one.

DeepSORT utilizes additional Re-ID models to enhance the long-range association. We surprisingly find BYTE also has additional gains compared with DeepSORT. This suggests a simple Kalman filter can perform long-range association and achieve better IDF1 and IDs when the detection boxes are accurate enough. We note that in severe occlusion cases, Re-ID features are vulnerable and may lead to identity switches, instead, motion model behaves more reliably.

\begin{table}[t]
\caption{Comparison of different data association methods on the MOT17 validation set. The best results are shown in \textbf{bold}. }
\begin{center}
{\input{tables/ass}}
\end{center}
\label{table_ass}
\vspace{-2mm}
\end{table}

MOTDT integrates motion-guided box propagation results along with detection results to associate unreliable detection results with tracklets. Although sharing the similar motivation, MOTDT is behind BYTE by a large margin. We explain that MOTDT uses propagated boxes as tracklet boxes, which may lead to locating drifts in tracking. Instead, BYTE uses low-score detection boxes to re-associate those unmatched tracklets, therefore, tracklet boxes are more accurate.

% Table~\ref{table_ass} also shows the results on BDD100K dataset. BYTE also outperforms other association methods by a large margin. Kalman filter fails in autonomous driving scenes and it is the main reason for the low performance of SORT, DeepSORT and MOTDT. Thus, we do not use Kalman filter on BDD100K. Additional off-the-shelf Re-ID models greatly improve the performance of BYTE on BDD100K. 

\begin{figure}[t]
	\centering
	\includegraphics[width=0.9\linewidth]{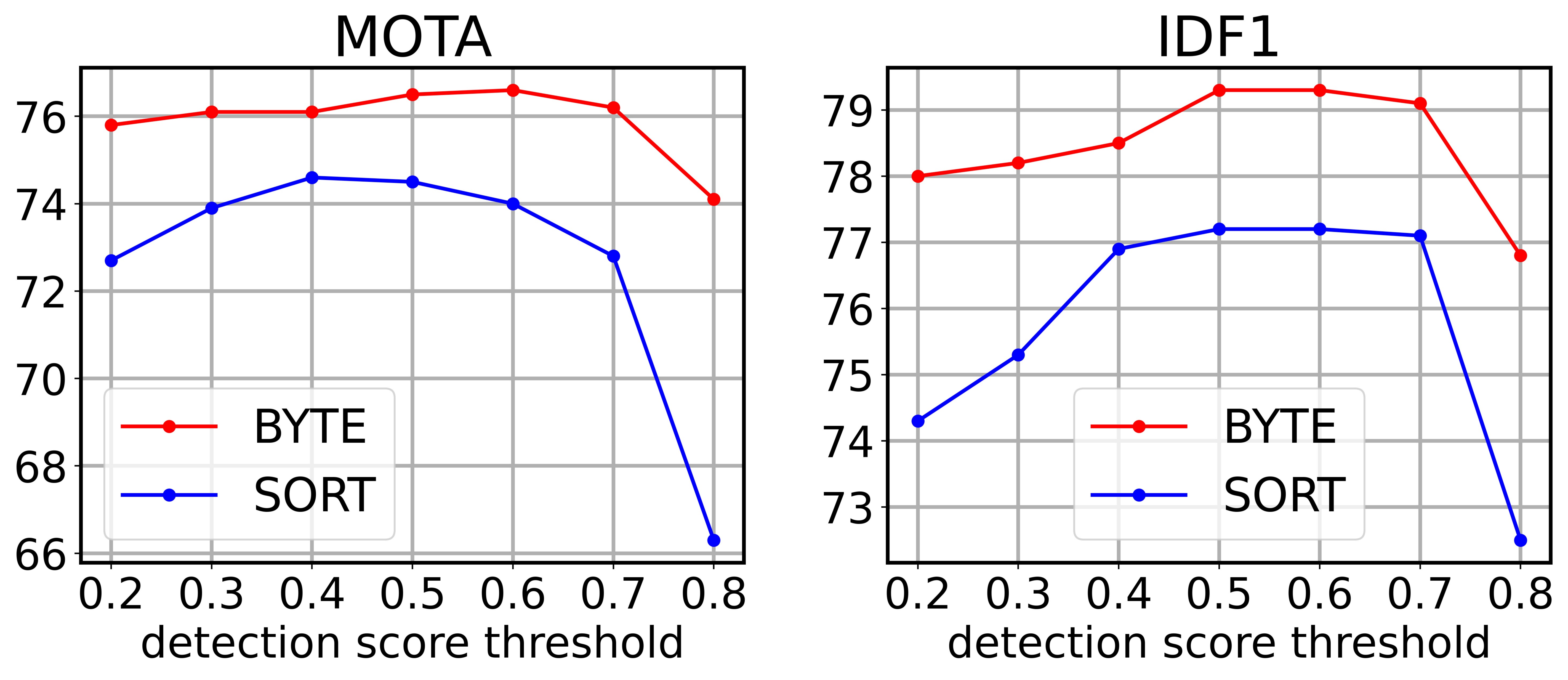}
	\caption{Comparison of the performances of BYTE and SORT under different detection score thresholds. The results are from the validation set of MOT17. }
	\label{fig:threshold}
	%\vspace{-3mm}
\end{figure}

\begin{figure*}[!htbp]
	\centering
	\includegraphics[width=0.8\linewidth]{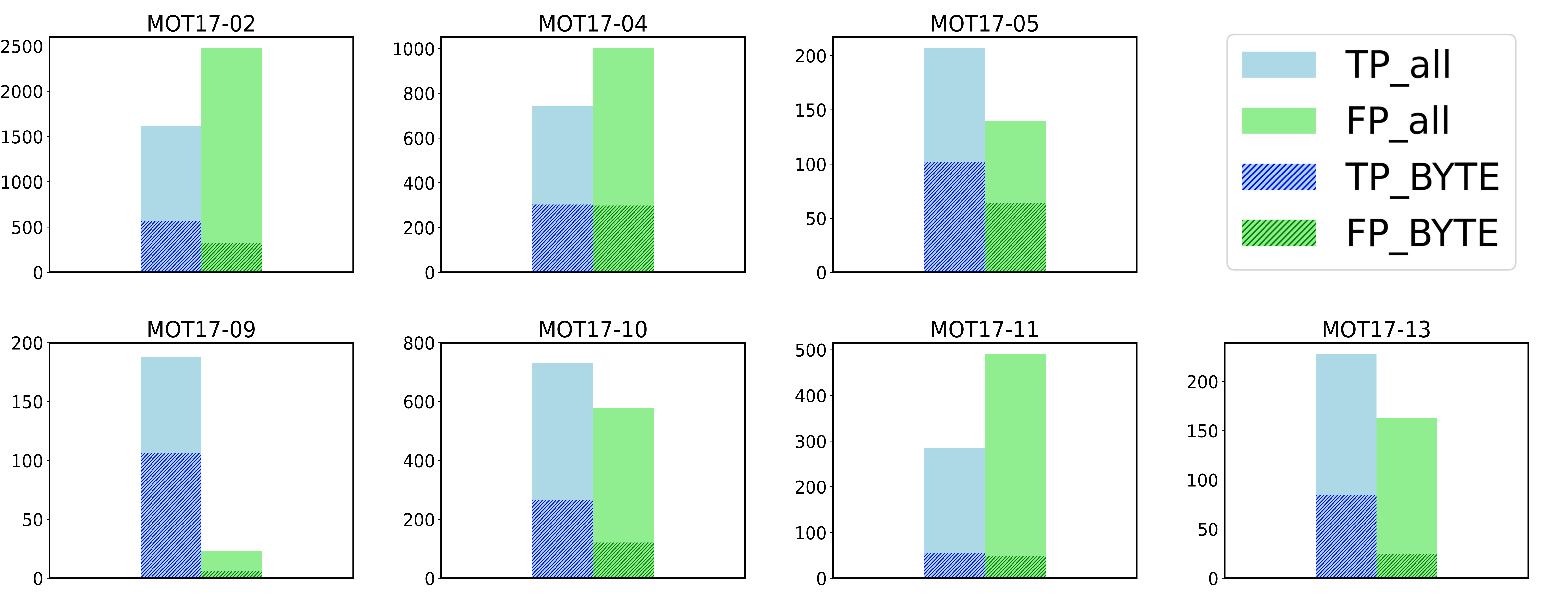}
	\caption{Comparison of the number of TPs and FPs in all low score detection boxes and the low score tracked boxes obtained by BYTE. The results are from the validation set of MOT17. }
	\label{fig:bar}
	\vspace{-3mm}
\end{figure*}

\myparagraph{Robustness to detection score threshold.}
The detection score threshold $\tau$ is a sensitive hyper-parameter and needs to be carefully tuned in the task of multi-object tracking. We change it from 0.2 to 0.8 and compare the MOTA and IDF1 score of BYTE and SORT. The results are shown in Fig.~\ref{fig:threshold}. From the results we can see that BYTE is more robust to the detection score threshold than SORT. This is because the second association in BYTE recovers the objects whose scores are lower than $\tau$, and thus considers almost every detection box regardless of the change of $\tau$.

% \begin{table}[htpb]
% \caption{Results of applying BYTE to 9 different SOTA trackers on the MOT17 validation set. ``K'' is short for Kalman filter. In green are the improvements of at least \textcolor{codegreen}{+\textbf{1.0}} point. }
% \begin{center}
% \scalebox{1.0}{\input{tables/app}}
% \end{center}
% \label{table_app}
% \vspace{-6mm}
% \end{table}

\myparagraph{Analysis on low score detection boxes.}
To prove the effectiveness of BYTE, we collect the number of TPs and FPs in the low score boxes obtained by BYTE. We use the half training set of MOT17 and CrowdHuman for training and evaluate on the half validation set of MOT17. First, we keep all the low score detection boxes whose scores range from $\tau_{low}$ to $\tau_{high}$ and classify the TPs and FPs using ground truth annotations. Then, we select the tracking results obtained by BYTE from low score detection boxes. The results of each sequence are shown in Fig.~\ref{fig:bar}. We can see that BYTE obtains notably more TPs than FPs from the low score detection boxes even though some sequences (\ie MOT17-02) have much more FPs in all the detection boxes. The obtained TPs notably increases MOTA from 74.6 to 76.6 as is shown in Table~\ref{table_ass}.

\begin{table}[t]
\caption{Comparison of the state-of-the-art methods under the “private detector” protocol on MOT17 test set. The best results are shown in \textbf{bold}. MOT17 contains rich scenes and half of the sequences are captured with camera motion. ByteTrack ranks 1st among all the trackers on the leaderboard of MOT17. It also has the highest running speed among all trackers.}
\begin{center}
\scalebox{1.0}{\input{tables/mot17}}
\end{center}
\label{table_mot17}
\vspace{-2mm}
\end{table}

\begin{table}[t]
\caption{Comparison of the state-of-the-art methods under the “private detector” protocol on MOT20 test set. The best results are shown in \textbf{bold}. The scenes in MOT20 are much more crowded than those in MOT17. ByteTrack ranks 1st among all the trackers on the leaderboard of MOT20. It also has the highest running speed among all trackers. }
\begin{center}
\scalebox{1.0}{\input{tables/mot20}}
\end{center}
\label{table_mot20}
\vspace{-3mm}
\end{table}

\begin{table}[t]
\caption{Comparison of the state-of-the-art methods under the “private detector” protocol on HiEve test set. The best results are shown in \textbf{bold}. HiEve has more complex events than MOT17 and MOT20. ByteTrack ranks 1st among all the trackers on the leaderboard of HiEve. }
\begin{center}
\scalebox{1.0}{\input{tables/hie}}
\end{center}
\label{table_hie}
\vspace{-5mm}
\end{table}

\subsection{Benchmark Evaluation}
We compare ByteTrack with the state-of-the-art trackers on the test set of MOT17, MOT20, HiEve and BDD100K under the private detection protocol in Table~\ref{table_mot17}, Table~\ref{table_mot20}, Table~\ref{table_hie} and Table~\ref{table_bdd} respectively. 
% We also evaluate ByteTrack on the test set of MOT17 and MOT20 under the public detection protocol in Table~\ref{table_mot17pub} and Table~\ref{table_mot20pub}. 

\myparagraph{MOT17.}
ByteTrack ranks 1st among all the trackers on the leaderboard of MOT17. Not only does it achieve the best accuracy (\ie 80.3 MOTA, 77.3 IDF1 and 63.1 HOTA), but also runs with highest running speed (30 FPS). It outperforms the second-performance tracker \cite{yang2021remot} by a large margin (\ie +3.3 MOTA, +5.3 IDF1 and +3.4 HOTA). Also, we use less training data than many high performance methods such as \cite{zhang2020fairmot,liang2020rethinking,wang2021multiple,liang2021one} (29K images vs. 73K images). It is worth noting that we only leverage the simplest similarity computation method Kalman filter in the association step compared to other methods \cite{zhang2020fairmot,liang2020rethinking,pang2021quasi,wang2020joint,zeng2021motr,sun2020transtrack} which additionally adopt Re-ID similarity or attention mechanisms. All these indicate that ByteTrack is a simple and strong tracker. 

%We also evaluate ByteTrack under the public detection protocol. Following the public detection filtering strategy in Tracktor \cite{bergmann2019tracking} and CenterTrack \cite{zhou2020tracking}, we only initialize a new trajectory when its IoU with a public detection box is larger than 0.8. We do not use tracklet interpolation under the public detection protocol. As is shown in Table~\ref{table_mot17pub}, ByteTrack outperforms other methods by a large margin on MOT17. For example, it outperforms SiamMOT by 1.5 points on MOTA and 6.7 points on IDF1. The results under public detection protocol further indicate the effectiveness of our association method by eliminating the effect of the detection performance. 

% \begin{table}[t]
% \caption{Comparison of the state-of-the-art methods under the “public detection” protocol on MOT17 test set. The best results are shown in \textbf{bold}. }
% \begin{center}
% \scalebox{1.0}{\input{tables/mot17_pub}}
% \end{center}
% \label{table_mot17pub}
% \vspace{-3mm}
% \end{table}

% \begin{table}[t]
% \caption{Comparison of the state-of-the-art methods under the “public detection” protocol on MOT20 test set. The best results are shown in \textbf{bold}. }
% \begin{center}
% \scalebox{1.0}{\input{tables/mot20_pub}}
% \end{center}
% \label{table_mot20pub}
% \vspace{-3mm}
% \end{table}

\myparagraph{MOT20.}
Compared with MOT17, MOT20 has much more crowded scenarios and occlusion cases. The average number of pedestrians in an image is 170 in the test set of MOT20. ByteTrack also ranks 1st among all the trackers on the leaderboard of MOT20 and outperforms other trackers by a large margin on almost all the metrics. For example, it increases MOTA from 68.6 to 77.8, IDF1 from 71.4 to 75.2 and decreases IDs by 71\% from 4209 to 1223. It is worth noting that ByteTrack achieves extremely low identity switches, which further indicates that associating every detection boxes is very effective under occlusion cases. 

%Table~\ref{table_mot20pub} shows the results under the public detection protocol. ByteTrack also outperforms existing results by a large margin. For example, it outperforms TMOH \cite{stadler2021improving} by 6.9 points on MOTA, 9.0 points on IDF1, 7.5 points on HOTA and reduce the identity switches by three quarters, which further indicates the effectiveness of our association method under crowded scenarios. 

\myparagraph{Human in Events.}
Compared with MOT17 and MOT20, HiEve contains more complex events and more diverse camera views. We train ByteTrack on CrowdHuman dataset and the training set of HiEve. ByteTrack also ranks 1st among all the trackers on the leaderboard of HiEve and outperforms other state-of-the-art trackers by a large margin. For example, it increases MOTA from 40.9 to 61.3 and IDF1 from 45.1 to 62.9. The superior results indicate that ByteTrack is robust to complex scenes. 

\begin{table}[t]
\caption{Comparison of the state-of-the-art methods on BDD100K test set. The best results are shown in \textbf{bold}. ByteTrack ranks 1st among all the trackers on the leaderboard of BDD100K. The methods denoted by * are the ones reported on the leaderboard of BDD100K. }
\begin{center}
\scalebox{1.0}{\input{tables/bdd100k}}
\end{center}
\label{table_bdd}
\vspace{-2mm}
\end{table}

\myparagraph{BDD100K.}
BDD100K is a multi-category tracking dataset in autonomous driving scenes. The challenges include low frame rate and large camera motion. We utilize a simple ResNet-50 ImageNet classification model from UniTrack \cite{wang2021different} to compute appearance similarity. ByteTrack ranks first on the leaderboard of BDD100K. It increases mMOTA from 36.6 to 45.5 on the validation set and 35.5 to 40.1 on the test set, which indicates that ByteTrack can also handle the challenges in autonomous driving scenes.

\begin{figure}[!h]
	\centering
	\includegraphics[width=0.9\linewidth]{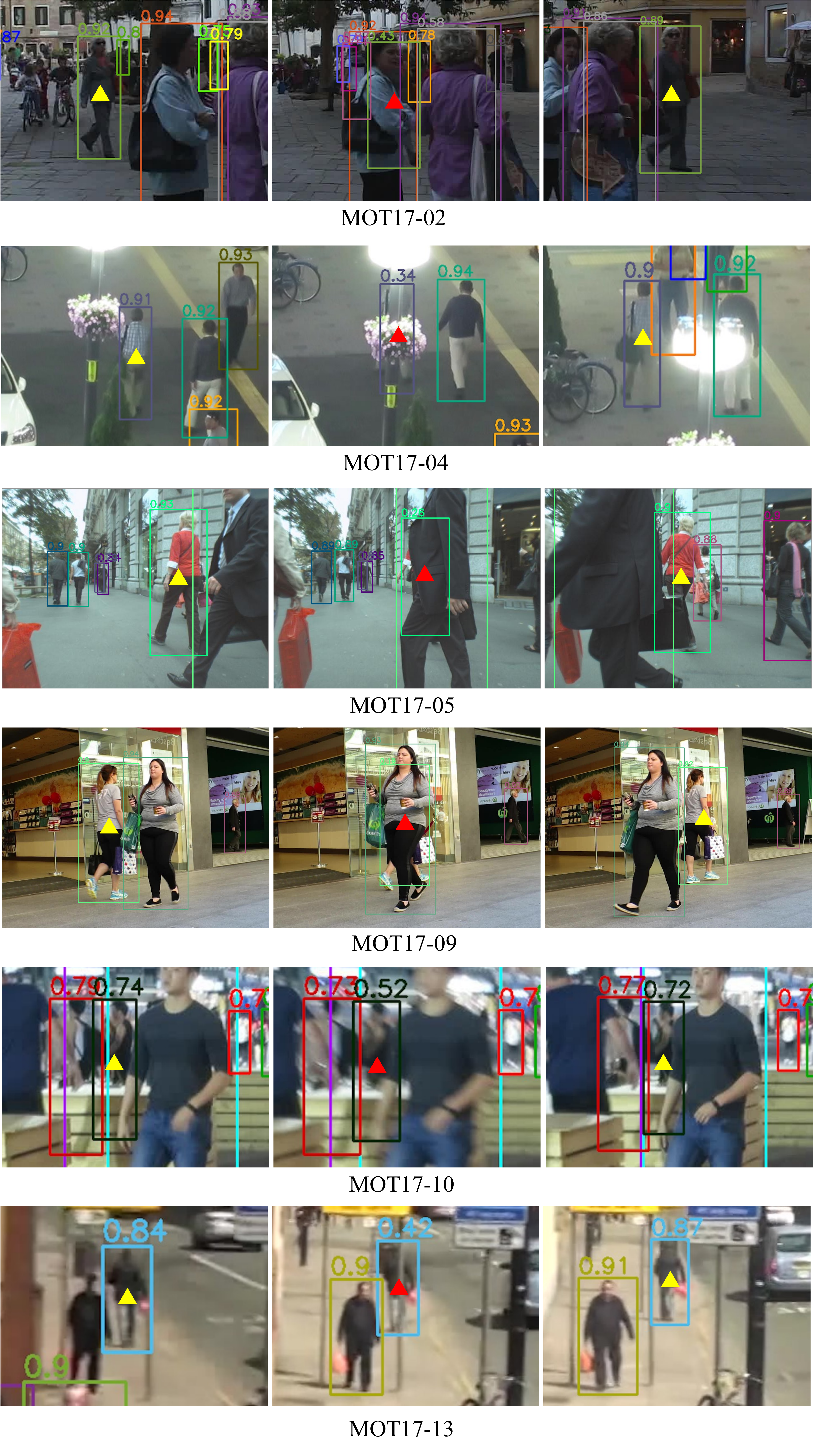}
	\caption{Visualization results of ByteTrack under the 2D MOT setting. We select 6 sequences from the validation set of MOT17 and show the effectiveness of ByteTrack to handle difficult cases such as occlusion and motion blur. The yellow triangle represents the high score box and the red triangle represents the low score box. The same box color represents the same identity. }
	\label{fig:vis}
	\vspace{-4mm}
\end{figure}

\myparagraph{Qualitative results. } We show some visualization results of difficult cases which ByteTrack is able to handle in Fig.~\ref{fig:vis}. We select 6 sequences from the half validation set of MOT17 and  generate the visualization results using the model with 76.6 MOTA and 79.3 IDF1. The difficult cases include occlusion (\ie MOT17-02, MOT17-04, MOT17-05, MOT17-09, MOT17-13), motion blur (\ie MOT17-10, MOT17-13) and small objects (\ie MOT17-13). The pedestrian in the middle frame with red triangle has low detection score, which is obtained by our association method. The low score boxes not only decrease the number of missing detections, but also play an important role for long-range association. As we can see from all these difficult cases, ByteTrack does not bring any identity switch and preserve the identity effectively. 

\subsection{3D MOT}
We evaluate \modelname under the 3D MOT setting in this section. We first present some experimental results of ablation analysis on the nuScenes dataset. Then, we compare \modelname with other trackers on the validation set and test set of nuScenes. All the experiments are conducted under both camera and LiDAR modalities to verify the generalization ability of \modelname. 

\subsubsection{Ablation Studies}
We present some results of ablation experiments on the design choices of each component such as complementary motion prediction and BYTE data association strategies in this section. We also discuss the effect of hyperparameters including the detection score threshold and the matching threshold. We adopt the detection results from PETRv2 \cite{liu2022petrv2} and CenterPoint \cite{yin2021center} under the camera modality and the LiDAR modality, respectively. 

\myparagraph{Complementary motion prediction. } As mentioned in Sec.~\ref{subsec:motion}, we propose a complementary motion prediction strategy consisting of the bilateral prediction of the Kalman filter and the detected velocity. We evaluate different motion prediction strategies in Table~\ref{table_motion_3d}. We can see that the detected velocity outperforms the Kalman filter motion model by 0.9 MOTA under the camera modality while the Kalman filter surpasses the detected velocity by 0.2 MOTA under the LiDAR modality. This is because LiDAR-based detectors provide more accurate detection results which help the Kalman filter obtain more reliable estimations. The complementary motion prediction achieves higher AMOTA and lower IDS than both the Kalman filter and the detected velocity, which indicates that the detected velocity is suitable for short-term association and the Kalman filter works for long-term association. When we further adaptively update the observation uncertainty matrix in Kalman filter based on the detection confidence following Eq.~\ref{eq:update}, the AMOTA metric is increased to 53.9 and 72.0 under the two modalities, which illustrates that detection helps motion prediction.

\myparagraph{Data association strategy. } We propose a detection-driven hierarchical data association strategy in Algorithm~\ref{algo:byte} of Sec.~\ref{subsec:asso}. We evaluate the effectiveness of the association strategy in Table~\ref{table_ass_3d}. For each modality, the first line marked with ``wo / BYTE'' in Table~\ref{table_ass_3d} means we only perform the high score detection boxes association in  Algorithm~\ref{algo:byte}. The second line marked with ``w / BYTE'' means we adopt the whole process of Algorithm~\ref{algo:byte} including both the high score and low score detection boxes associations. We can see that the low score detection boxes association boosts the 3D tracking performance by 0.3 AMOTA under the camera modality and 0.4 AMOTA under the LiDAR modality. The experimental results illustrate that the detection-driven hierarchical data association strategy is effective under both 2D and 3D settings by mining true objects in the low-score detection boxes.

\begin{figure}[!h]
	\centering
	\includegraphics[width=0.98\linewidth]{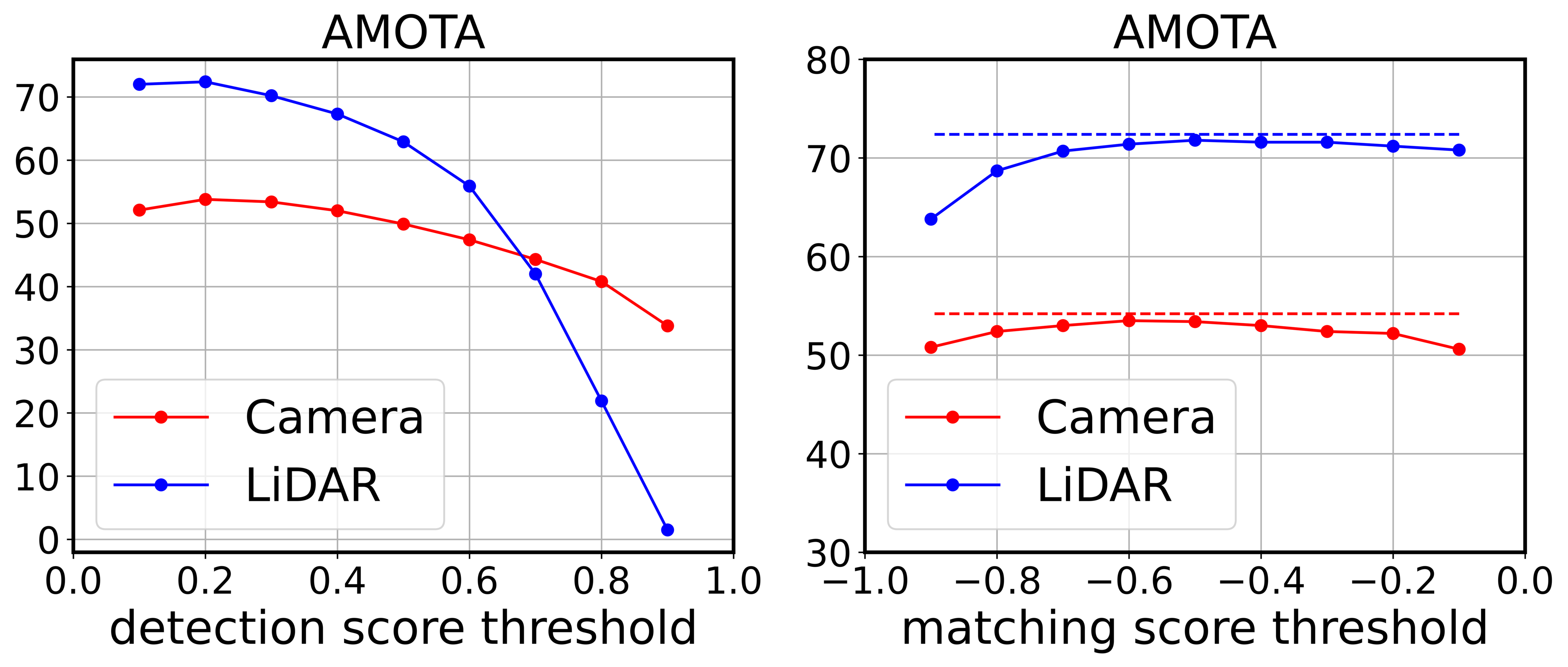}
	\caption{Comparison of the tracking performance under different detection score thresholds and matching score thresholds. The dashed line represents using the optimal matching score threshold for each class. }
	\label{fig:hyper}
% 	\vspace{-2mm}
\end{figure}

\myparagraph{Hyperparameter search. } The detection score threshold and the GIoU matching score threshold are two important hyperparameters in the entire tracking framework. From the left part of Fig.~\ref{fig:hyper} we can see that a lower detection score threshold tends to achieve higher AMOTA as AMOTA needs high recall of objects. However, simply decreasing the detection score threshold may bring a large number of wrong associations which harm the tracking performance. We perform grid search and find that the optimal detection score threshold is 0.25 for camera-based detectors and 0.2 for LiDAR-based detectors. The right part of Fig.~\ref{fig:hyper} shows how the matching score threshold affects AMOTA. We find that the optimal GIoU threshold is around -0.5 under both the LiDAR and Camera settings. We also observe that objects of different classes need different matching score thresholds because they have different sizes and velocities. The dashed line shows the performance of using optimal thresholds for each class, which brings about 0.5 AMOTA improvements. The optimal GIoU threshold for each class is listed in Sec.~\ref{subsec:detail}. 

\begin{table}[t]
\caption{Comparison of different motion prediction strategies on nuScenes validation set. We evaluate using both camera and LiDAR modalities. ``DV'' is short of detected velocity. }
\begin{center}
\scalebox{1.0}{\input{tables/motion_3d}}
\end{center}
\label{table_motion_3d}
\vspace{-4mm}
\end{table}

\begin{table}[t]
\caption{Comparison of different data association strategies on the validation set of nuScenes using both camera and LiDAR modalities. }
\begin{center}
\scalebox{1.0}{\input{tables/ass_3d}}
\end{center}
\label{table_ass_3d}
\vspace{-4mm}
\end{table}

\subsubsection{Benchmark Evaluation}
We compare \modelname with the state-of-the-art trackers on the validation set and test set of nuScenes under both camera and LiDAR modalities in Table~\ref{table_nu_camera} and Table~\ref{table_nu_lidar} respectively. 

\myparagraph{Camera modality. } We directly adopt the detection results from PETRv2 \cite{liu2022petrv2} as input. \modelname ranks 1st among all the camera-based methods on both the validation set and the test set of the nuScenes dataset. As is shown in Table~\ref{table_nu_camera}, it achieves 54.2 AMOTA and 696 IDS on the validation set, which surpasses the second one QTrack \cite{yang2022quality} by 3.1 AMOTA and half the number of identity switches. On the test set, it greatly increases AMOTA from 51.9 to 56.4 and decreases IDS by 68\% from 2204 to 704. The low number of identity switches indicates the effectiveness of our tracking algorithm. It also outperforms some LiDAR-based methods such as \cite{weng20203d,chiu2021pro} in Table~\ref{table_nu_lidar}, which significantly narrows the performance gap between camera-based and LiDAR-based methods.

\begin{table}[t]
\caption{Comparison of the state-of-the-art camera-based methods on the nuScenes dataset. The best results are shown in \textbf{bold}. \modelname ranks 1st among all the camera-based trackers on the leaderboard of nuScenes. }
\begin{center}
\scalebox{1.0}{\input{tables/nu_camera}}
\end{center}
\label{table_nu_camera}
\vspace{-3mm}
\end{table}

\myparagraph{LiDAR modality. } We utilize the publicly available CenterPoint \cite{yin2021center} and TransFusion-L \cite{bai2022transfusion} detection results as input on the validation set of nuScenes. For fair comparisons, we list the results using the same CenterPoint detection as all other methods \cite{weng2020ab3dmot,chiu2021pro,yin2021center,benbarka2021score,zaech2022learnable,pang2021simpletrack,wang2021immortal} on the validation set. From Table~\ref{table_nu_lidar} we can see that \modelname* outperforms the second one Immortal \cite{wang2021immortal} by 2.2 AMOTA, 2.3 MOTA and 52\% IDS using the same detection results, which illustrates the advantages of our data association and motion prediction strategies. When equipped with powerful TransFusion-L detection results, it further increases the AMOTA metric to 75.0 and surpasses other methods by a large margin. On the test set of nuScenes, we adopt the same detection results as TransFusion-L. \modelname outperforms TransFusion-L by 1.5 AMOTA and 45\% IDS. The improvements come entirely from our tracking algorithm. \modelname is the first LiDAR-based method which achieves 70 AMOTA and set the new state-of-the-art result on nuScenes. 

\begin{table}[t]
\caption{Comparison of the state-of-the-art LiDAR-based methods on the nuScenes dataset. The best results are shown in \textbf{bold}. \modelname ranks 1st among all the LiDAR-based trackers on the leaderboard of nuScenes. * represents that we use the same CenterPoint \cite{yin2021center} detection as all other methods on the validation set. }
\begin{center}
\scalebox{1.0}{\input{tables/nu_lidar}}
\end{center}
\label{table_nu_lidar}
\vspace{-0mm}
\end{table}

\begin{figure}[!h]
	\centering
	\includegraphics[width=0.9\linewidth]{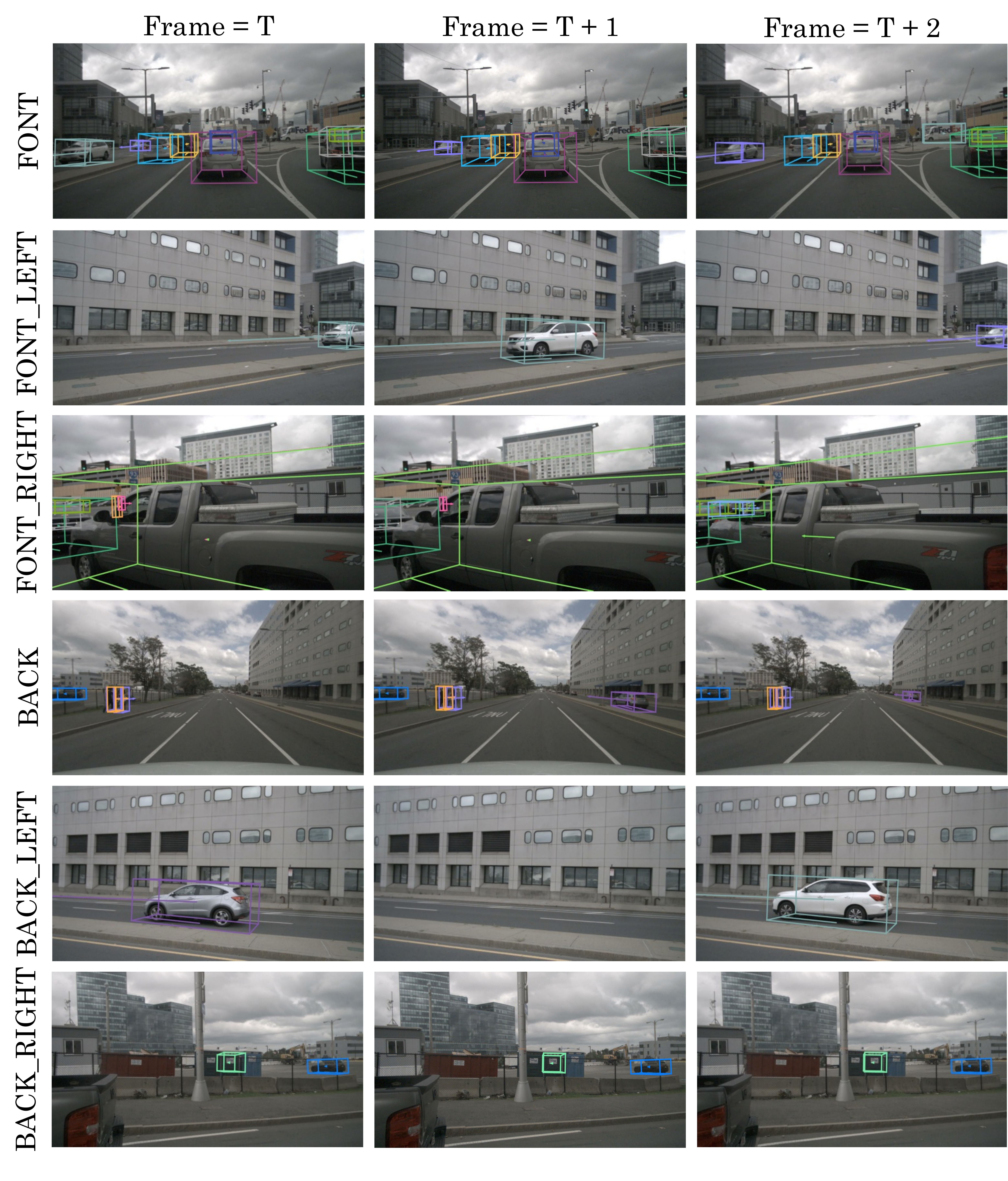}
	\caption{Visualization results of \modelname under the camera 3D MOT setting. We select one sequence from nuScenes validation set and show the results from 6 different cameras. The same box color represents the same identity. The arrow represents the direction and magnitude of the object velocity. }
	\label{fig:cam_vis}
	\vspace{-0mm}
\end{figure}

\begin{figure}[!h]
	\centering
	\includegraphics[width=0.9\linewidth]{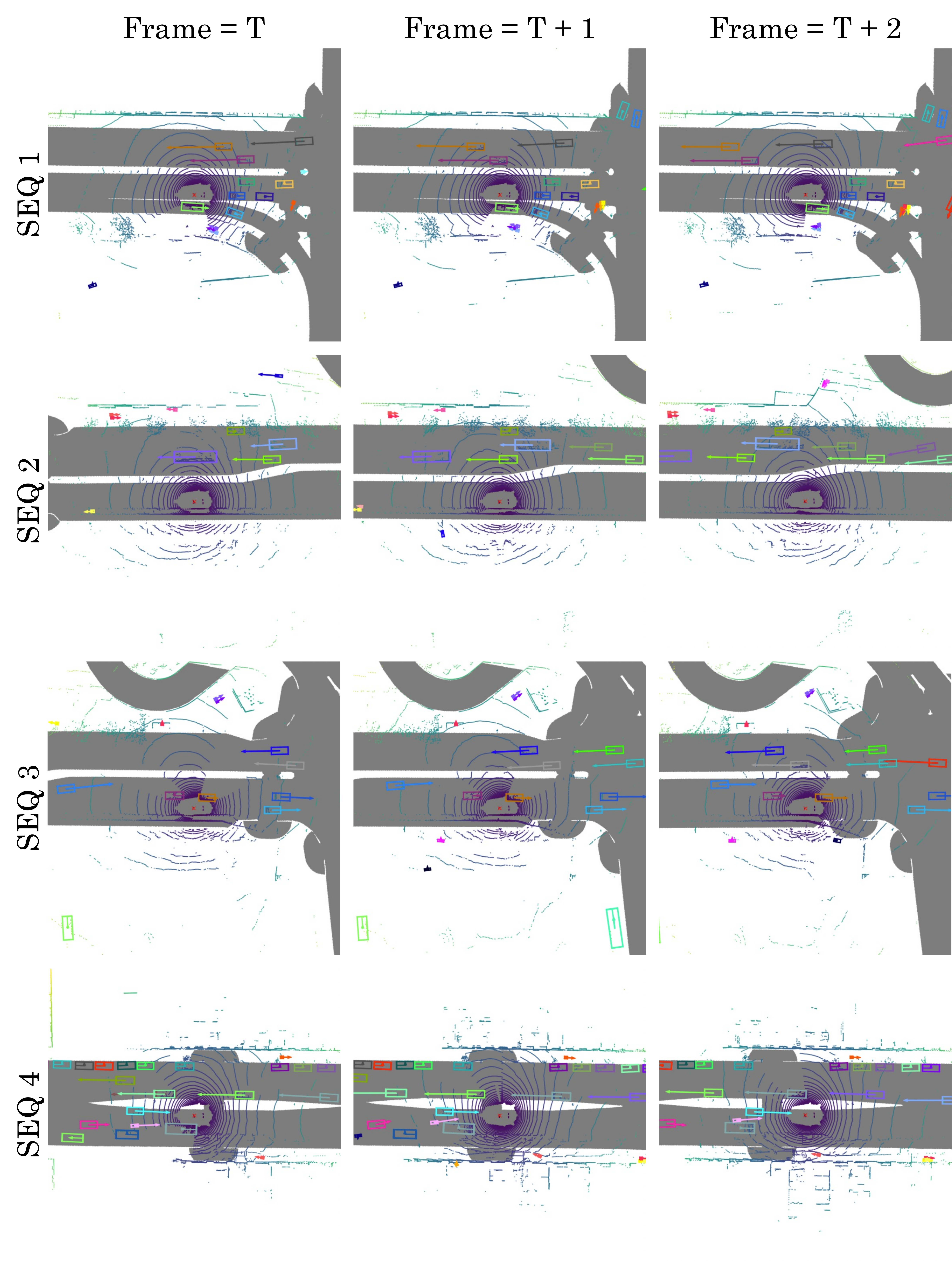}
	\caption{Visualization results of \modelname under the LiDAR 3D MOT setting. We select 4 sequences from nuScenes validation set. Colors and arrows have similar meanings with those under the camera setting. }
	\label{fig:lidar_vis}
	\vspace{-0mm}
\end{figure}

\myparagraph{Qualitative results. } We show some visualization results of difficult cases which \modelname is able to handle under the camera and LiDAR 3D MOT settings in Fig.~\ref{fig:cam_vis} and Fig.~\ref{fig:lidar_vis} respectively. The difficult cases include occlusion (\ie FONT and FONT\_RIGHT cameras in Fig.~\ref{fig:cam_vis}, SEQ 4 in Fig.~\ref{fig:lidar_vis}) and abrupt motion (\ie FONT\_LEFT camera in Fig.~\ref{fig:cam_vis}, all the sequences in Fig.~\ref{fig:lidar_vis}).  As we can see from all these difficult cases, \modelname does not bring any identity switch and preserve the identity effectively.

%% file: tables/sim.tex
% \setlength{\tabcolsep}{3.0pt}

% \begin{tabular}{ l l c c c c c c}

% \toprule
%  & & & MOT17 & & & BDD100K & \\
% Sim\#1 & Sim\#2 & MOTA$\uparrow$ & IDF1$\uparrow$ & IDs$\downarrow$ & mMOTA$\uparrow$ & mIDF1$\uparrow$ & IDs$\downarrow$\\
% % \midrule
% \cmidrule(lr){1-2} \cmidrule(lr){3-5} \cmidrule(lr){6-8}
% IoU & Re-ID & 75.8 & 77.5 & 231 & 39.2 & 48.3 & 29172\\
% IoU & IoU & \textbf{76.6} & 79.3 & \textbf{159} & 39.4 & 48.9 & 27902\\
% Re-ID & Re-ID & 75.2 & 78.7 & 276 & 45.0 & 53.4 & 10425\\
% Re-ID & IoU & 76.3 & \textbf{80.5} & 216 & \textbf{45.5} & \textbf{54.8} & \textbf{9140}\\
% \bottomrule
% \end{tabular}

\setlength{\tabcolsep}{12pt}

\begin{tabular}{ l l c c c}

\toprule
Sim\#1 & Sim\#2 & MOTA$\uparrow$ & IDF1$\uparrow$ & IDs$\downarrow$ \\
\midrule
IoU & Re-ID & 75.8 & 77.5 & 231 \\
\cellcolor{Gray}IoU & \cellcolor{Gray}IoU & \cellcolor{Gray}\textbf{76.6} & \cellcolor{Gray}79.3 & \cellcolor{Gray}\textbf{159}\\
Re-ID & Re-ID & 75.2 & 78.7 & 276 \\
Re-ID & IoU & 76.3 & \textbf{80.5} & 216 \\
\bottomrule
\end{tabular}

%% file: tables/ass.tex
\setlength{\tabcolsep}{6pt}
\begin{tabular}{ l c c c c c}
\toprule
Method & w/ Re-ID & MOTA$\uparrow$ & IDF1$\uparrow$ & IDs$\downarrow$ & FPS \\
\midrule
SORT &  & 74.6 & 76.9 & 291 & \textbf{30.1}\\
DeepSORT & $\checkmark$ & 75.4 & 77.2 & 239 & 13.5\\
MOTDT & $\checkmark$ & 75.8 & 77.6 & 273 & 11.1\\
\cellcolor{Gray}BYTE & \cellcolor{Gray} & \cellcolor{Gray}\textbf{76.6} & \cellcolor{Gray}\textbf{79.3} & \cellcolor{Gray}\textbf{159} & \cellcolor{Gray}29.6 \\
\bottomrule
\end{tabular}

% \setlength{\tabcolsep}{1.0pt}
% \begin{tabular}{ l c c c c c c c}
% \toprule
% & & & MOT17 & &  & BDD100K & \\
% Method & w/ ReID & MOTA$\uparrow$ & IDF1$\uparrow$ & IDs$\downarrow$ & mMOTA$\uparrow$ & mIDF1$\uparrow$ & IDs$\downarrow$ \\
% % \midrule
% \cmidrule(lr){1-2} \cmidrule(lr){3-5} \cmidrule(lr){6-8} 
% SORT &  & 74.6 & 76.9 & 291 & 30.9 & 41.3 & 10067 \\
% DeepSORT & $\checkmark$ & 75.4 & 77.2 & 239 & 24.5 & 38.2 & 10720 \\
% MOTDT & $\checkmark$ & 75.8 & 77.6 & 273 & 26.7 & 39.8 & 14520 \\
% \cellcolor{Gray}BYTE & \cellcolor{Gray} & \cellcolor{Gray}\textbf{76.6} & \cellcolor{Gray}79.3 & \cellcolor{Gray}\textbf{159} & \cellcolor{Gray}39.4 & \cellcolor{Gray}48.9 & \cellcolor{Gray}27902 \\
% \cellcolor{Gray}BYTE & \cellcolor{Gray}$\checkmark$ & \cellcolor{Gray}76.3 & \cellcolor{Gray}\textbf{80.5} & \cellcolor{Gray}216 & \cellcolor{Gray}\textbf{45.5} & \cellcolor{Gray}\textbf{54.8} & \cellcolor{Gray}\textbf{9140} \\
% \bottomrule
% \end{tabular}

%% file: tables/mot17.tex
\setlength{\tabcolsep}{1.0pt}

\begin{tabular}{ l c c c c c c c}

\toprule
Tracker & MOTA$\uparrow$ & IDF1$\uparrow$ & HOTA$\uparrow$ & FP$\downarrow$ & FN$\downarrow$ & IDs$\downarrow$ & FPS$\uparrow$\\
\midrule
%DAN \cite{sun2019deep} & 52.4 & 49.5 & 39.3 & 25423 & 234592 & 8431 & \textless 3.9\\ 
Tube\_TK \cite{pang2020tubetk} & 63.0 & 58.6 & 48.0 & 27060 & 177483 & 4137 & 3.0\\
MOTR \cite{zeng2021motr} & 65.1 & 66.4 & - & 45486 & 149307 & \textbf{2049} & -\\
CTracker \cite{peng2020chained} & 66.6 & 57.4 & 49.0 & 22284 & 160491 & 5529 & 6.8\\
CenterTrack \cite{zhou2020tracking} & 67.8 & 64.7 & 52.2 & \textbf{18498} & 160332 & 3039 & 17.5\\
QuasiDense \cite{pang2021quasi} & 68.7 & 66.3 & 53.9 & 26589 & 146643 & 3378 & 20.3\\
TraDes \cite{wu2021track} & 69.1 & 63.9 & 52.7 & 20892 & 150060 & 3555 & 17.5\\
SOTMOT \cite{zheng2021improving} & 71.0 & 71.9 & - & 39537 & 118983 & 5184 & 16.0\\
TransCenter \cite{xu2021transcenter} & 73.2 & 62.2 & 54.5 & 23112 & 123738 & 4614 & 1.0\\
GSDT \cite{wang2020joint} & 73.2 & 66.5 & 55.2 & 26397 & 120666 & 3891 & 4.9\\
FairMOT \cite{zhang2020fairmot} & 73.7 & 72.3 & 59.3 & 27507 & 117477 & 3303 & 25.9\\
PermaTrackPr \cite{tokmakov2021learning} & 73.8 & 68.9 & 55.5 & 28998 & 115104 & 3699 & 11.9\\
CSTrack \cite{liang2020rethinking} & 74.9 & 72.6 & 59.3 & 23847 & 114303 & 3567 & 15.8\\
TransTrack \cite{sun2020transtrack} & 75.2 & 63.5 & 54.1 & 50157 & 86442 & 3603 & 10.0\\
SiamMOT \cite{liang2021one} & 76.3 & 72.3 & - & - & - & - & 12.8\\
CorrTracker \cite{wang2021multiple} & 76.5 & 73.6 & 60.7 & 29808 & 99510 & 3369 & 15.6\\
TransMOT \cite{chu2021transmot} & 76.7 & 75.1 & 61.7 & 36231 & 93150 & 2346 & 9.6\\
ReMOT \cite{yang2021remot} & 77.0 & 72.0 & 59.7 & 33204 & 93612 & 2853 & 1.8\\
\cellcolor{Gray}ByteTrack & \cellcolor{Gray}\textbf{80.3} & \cellcolor{Gray}\textbf{77.3} & \cellcolor{Gray}\textbf{63.1} & \cellcolor{Gray}25491 & \cellcolor{Gray}\textbf{83721} & \cellcolor{Gray}2196 & \cellcolor{Gray}\textbf{29.6}\\
\bottomrule
\end{tabular}

%% file: tables/mot20.tex
\setlength{\tabcolsep}{1.0pt}

\begin{tabular}{ l c c c c c c c}

\toprule
Tracker & MOTA$\uparrow$ & IDF1$\uparrow$ & HOTA$\uparrow$ & FP$\downarrow$ & FN$\downarrow$ & IDs$\downarrow$ & FPS$\uparrow$\\
\midrule
FairMOT \cite{zhang2020fairmot} & 61.8 & 67.3 & 54.6 & 103440 & 88901 & 5243 & 13.2\\
TransCenter \cite{xu2021transcenter} & 61.9 & 50.4 & - & 45895 & 146347 & 4653 & 1.0\\
TransTrack \cite{sun2020transtrack} & 65.0 & 59.4 & 48.5 & 27197 & 150197 & 3608 & 7.2\\
CorrTracker \cite{wang2021multiple} & 65.2 & 69.1 & - & 79429 & 95855 & 5183 & 8.5\\
CSTrack \cite{liang2020rethinking} & 66.6 & 68.6 & 54.0 & \textbf{25404} & 144358 & 3196 & 4.5\\
GSDT \cite{wang2020joint} & 67.1 & 67.5 & 53.6 & 31913 & 135409 & 3131 & 0.9\\
SiamMOT \cite{liang2021one} & 67.1 & 69.1 & - & - & - & - & 4.3\\
SOTMOT \cite{zheng2021improving} & 68.6 & 71.4 & - & 57064 & 101154 & 4209 & 8.5\\
\cellcolor{Gray}ByteTrack & \cellcolor{Gray}\textbf{77.8} & \cellcolor{Gray}\textbf{75.2} & \cellcolor{Gray}\textbf{61.3} & \cellcolor{Gray}26249 & \cellcolor{Gray}\textbf{87594} & \cellcolor{Gray}\textbf{1223} & \cellcolor{Gray}\textbf{17.5}\\
\bottomrule
\end{tabular}

%% file: tables/hie.tex
\setlength{\tabcolsep}{2.0pt}

\begin{tabular}{ l c c c c c c c}

\toprule
Tracker & MOTA$\uparrow$ & IDF1$\uparrow$ & MT$\uparrow$ & ML$\downarrow$ & FP$\downarrow$ & FN$\downarrow$ & IDs$\downarrow$\\
\midrule
DeepSORT \cite{wojke2017simple} & 27.1 & 28.6 & 8.5\% & 41.5\% & 5894 & 42668 & 2220\\
MOTDT \cite{chen2018real} & 26.1 & 32.9 & 8.7\% & 54.6\% & 6318 & 43577 & 1599\\
IOUtracker \cite{bochinski2017high} & 38.6 & 38.6 & 28.3\% & 27.6\% & 9640 & 28993 & 4153\\
JDE \cite{wang2020towards} & 33.1 & 36.0 & 15.1\% & 24.1\% & 9526 & 33327 & 3747\\
FairMOT \cite{zhang2020fairmot} & 35.0 & 46.7 & 16.3\% & 44.2\% & 6523 & 37750 & \textbf{995}\\
CenterTrack \cite{zhou2020tracking} & 40.9 & 45.1 & 10.8\% & 32.2\% & 3208 & 36414 & 1568\\
\cellcolor{Gray}ByteTrack & \cellcolor{Gray}\textbf{61.7} & \cellcolor{Gray}\textbf{63.1} & \cellcolor{Gray}\textbf{38.3\%} & \cellcolor{Gray}\textbf{21.6\%} & \cellcolor{Gray}\textbf{2822} & \cellcolor{Gray}\textbf{22852} & \cellcolor{Gray}1031\\
\bottomrule
\end{tabular}

%% file: tables/bdd100k.tex
% \setlength{\tabcolsep}{0.8pt}

\setlength{\tabcolsep}{0.5pt}

\begin{tabular}{l cccccccc}
\toprule
Tracker & Split & mMOTA$\uparrow$ & mIDF1$\uparrow$ & MOTA$\uparrow$ & IDF1$\uparrow$ & FN$\downarrow$ & FP$\downarrow$ & IDs$\downarrow$ \\
\midrule
Yu \etal \cite{yu2020bdd100k} & val & 25.9 & 44.5 & 56.9 & 66.8 & 122406 & 52372 & 8315 \\
QDTrack \cite{pang2021quasi} & val & 36.6 & 50.8 & 63.5 & \textbf{71.5} & 108614 & 46621 & \textbf{6262} \\
\cellcolor{Gray}ByteTrack & \cellcolor{Gray}val & \cellcolor{Gray}\textbf{45.5} & \cellcolor{Gray}\textbf{54.8} & \cellcolor{Gray}\textbf{69.1} & \cellcolor{Gray}70.4 & \cellcolor{Gray}\textbf{92805} & \cellcolor{Gray}\textbf{34998} & \cellcolor{Gray}9140 \\
\midrule
Yu \etal \cite{yu2020bdd100k} & test & 26.3 & 44.7 & 58.3 & 68.2 & 213220 & 100230 & 14674 \\

DeepBlueAI* & test & 31.6 & 38.7 & 56.9& 56.0& 292063 & \textbf{35401} & 25186 \\

Madamada* & test & 33.6 & 43.0 & 59.8& 55.7& 209339 & 76612  & 42901 \\

QDTrack\cite{pang2021quasi} & test & 35.5 & 52.3 & 64.3 & \textbf{72.3}  & 201041 & 80054  & \textbf{10790} \\

\cellcolor{Gray}ByteTrack & \cellcolor{Gray}test & \cellcolor{Gray}\textbf{40.1} & \cellcolor{Gray}\textbf{55.8} & \cellcolor{Gray}\textbf{69.6} & \cellcolor{Gray}71.3 & \cellcolor{Gray}\textbf{169073} & \cellcolor{Gray}63869 & \cellcolor{Gray}15466 \\
\bottomrule
\end{tabular}

%% file: tables/motion_3d.tex
% \setlength{\tabcolsep}{5.0pt}

% \begin{tabular}{ l c c c c}

% \toprule
% Motion Prediction & Modality & AMOTA\%$\uparrow$ & MOTA\%$\uparrow$ & IDS$\downarrow$ \\
% \midrule
% Kalman & Camera & 52.2 & 45.0 & 1149 \\
% PV & Camera & 53.1 & 45.9 & 839 \\
% Integrated & Camera & 53.7 & 46.8 & 753 \\
% Integrated + Update & Camera & \textbf{53.9} & \textbf{46.8} & \textbf{747} \\
% \midrule
% Kalman & LiDAR & 70.8 & 61.2 & 284 \\
% PV & LiDAR &  &  &  \\
% Integrated & LiDAR & 71.3 & 62.0 & 201 \\
% Integrated + Update & LiDAR & \textbf{72.0} & \textbf{62.7} & \textbf{179} \\

% \bottomrule
% \end{tabular}

\setlength{\tabcolsep}{10.0pt}

\begin{tabular}{ l c c c}

\toprule
Motion Prediction & Modality & AMOTA\%$\uparrow$ & IDS$\downarrow$ \\
\midrule
Kalman & Camera & 52.2 & 1149 \\
DV & Camera & 53.1 & 839 \\
Integrated & Camera & 53.7 & 753 \\
Integrated + Update & Camera & \textbf{53.9} & \textbf{747} \\
\midrule
Kalman & LiDAR & 70.8 & 284 \\
DV & LiDAR & 70.6 & 231 \\
Integrated & LiDAR & 71.3 & 201 \\
Integrated + Update & LiDAR & \textbf{72.0} & \textbf{179} \\

\bottomrule
\end{tabular}

%% file: tables/ass_3d.tex
\setlength{\tabcolsep}{10.0pt}

\begin{tabular}{ l c c c}

\toprule
Data Association & Modality & AMOTA\%$\uparrow$ & IDS$\downarrow$ \\
\midrule
wo / BYTE & Camera & 53.9 & 747 \\
w / BYTE & Camera & \textbf{54.2} & \textbf{696} \\
\midrule
wo / BYTE & LiDAR & 72.0 & \textbf{179} \\
w / BYTE & LiDAR & \textbf{72.4} & 183 \\

\bottomrule
\end{tabular}

%% file: tables/nu_camera.tex
% \setlength{\tabcolsep}{2.0pt}

\setlength{\tabcolsep}{3.0pt}

\begin{tabular}{ l c c c c c}

\toprule
Methods & Split & AMOTA\%$\uparrow$ & AMOTP(m)$\downarrow$ & MOTA\%$\uparrow$ & IDS$\downarrow$ \\
\midrule
DEFT \cite{chaabane2021deft} & val & 20.1 & - & 17.1 & - \\
QD3DT \cite{hu2022monocular} & val & 24.2 & 1.518 & 21.8 & 5646 \\
TripletTrack \cite{marinello2022triplettrack} & val & 28.5 & 1.485 & - & - \\
MUTR3D \cite{zhang2022mutr3d} & val & 29.4 & 1.498 & 26.7 & 3822 \\
QTrack \cite{yang2022quality} & val & 51.1 & \textbf{1.090} & \textbf{46.5} & 1144\\
\cellcolor{Gray}\modelname & \cellcolor{Gray}val & \cellcolor{Gray}\textbf{54.2} & \cellcolor{Gray}1.108 & \cellcolor{Gray}\textbf{46.5} & \cellcolor{Gray}\textbf{696}\\
\midrule
CenterTrack \cite{zhou2020tracking} & test & 4.6 & 1.543 & 4.3 & 3807\\
DEFT \cite{chaabane2021deft} & test & 17.7 & 1.564 & 15.6 & 6901 \\
Time3D \cite{li2022time3d} & test & 21.0 & 1.360 & 17.3 & - \\
QD3DT \cite{hu2022monocular} & test & 21.7 & 1.550 & 19.8 & 6856 \\
TripletTrack \cite{marinello2022triplettrack} & test & 26.8 & 1.504 & 24.5 & 1044 \\
MUTR3D \cite{zhang2022mutr3d} & test & 27.0 & 1.494 & 24.5 & 6018 \\
PolarDETR \cite{chen2022polar} & test & 27.3 & 1.185 & 23.8 & 2170\\
SRCN3D \cite{shi2022srcn3d} & test & 39.8 & 1.317 & 35.9 & 4090\\
QTrack \cite{yang2022quality} & test & 48.0 & 1.107 & 43.1 & 1484\\
UVTR \cite{li2022unifying} & test & 51.9 & 1.125 & 44.7 & 2204\\
\cellcolor{Gray}\modelname & \cellcolor{Gray}test & \cellcolor{Gray}\textbf{56.4} & \cellcolor{Gray}\textbf{1.005} & \cellcolor{Gray}\textbf{47.1} & \cellcolor{Gray}\textbf{704}\\

\bottomrule
\end{tabular}

%% file: tables/nu_lidar.tex
\setlength{\tabcolsep}{3.0pt}

\begin{tabular}{ l c c c c c}

\toprule
Methods & Split & AMOTA\%$\uparrow$ & AMOTP(m)$\downarrow$ & MOTA\%$\uparrow$ & IDS$\downarrow$ \\
\midrule
AB3DMOT \cite{weng20203d} & val & 57.8 & 0.807 & 51.4 & 1275 \\
Chiu \etal \cite{chiu2021pro} & val & 61.7 & 0.984 & 53.3 & 680 \\
CenterPoint \cite{yin2021center} & val & 66.5 & 0.567 & 56.2 & 562 \\
CBMOT \cite{benbarka2021score} & val & 67.5 & 0.591 & 58.3 & 494 \\
OGR3MOT \cite{zaech2022learnable} & val & 69.3 & 0.627 & 60.2 & 262\\
SimpleTrack \cite{pang2021simpletrack} & val & 69.6 & 0.547 & 60.2 & 405\\
Immortal \cite{wang2021immortal} & val & 70.2 & \textbf{0.524} & 60.1 & 385\\
\cellcolor{Gray}\modelname* & \cellcolor{Gray}val & \cellcolor{Gray}72.4 & \cellcolor{Gray}0.547 & \cellcolor{Gray}62.4 & \cellcolor{Gray}\textbf{183}\\
\cellcolor{Gray}\modelname & \cellcolor{Gray}val & \cellcolor{Gray}\textbf{75.0} & \cellcolor{Gray}0.534 & \cellcolor{Gray}\textbf{64.7} & \cellcolor{Gray}367\\
\midrule
AB3DMOT \cite{weng20203d} & test & 15.1 & 1.501 & 15.4 & 9027\\
Chiu \etal \cite{chiu2021pro} & test & 55.0 & 0.798 & 45.9 & 776 \\
CenterPoint \cite{yin2021center} & test & 63.8 & 0.555 & 53.7 & 760 \\
CBMOT \cite{benbarka2021score} & test & 64.9 & 0.592 & 54.5 & 557\\
PolarMOT \cite{kim2022polarmot} & test & 66.4 & 0.566 & 56.1 & 242\\
OGR3MOT \cite{zaech2022learnable} & test & 65.6 & 0.620 & 55.4 & 288\\
BP \cite{meyer2018message} & test & 66.6 & 0.571 & 57.1 & \textbf{182}\\
SimpleTrack \cite{pang2021simpletrack} & test & 66.8 & 0.550 & 56.6 & 575\\
UVTR \cite{li2022unifying} & test & 67.0 & 0.656 & 56.1 & 774\\
% ShaSTA \cite{sadjadpour2022shasta} & test & 67.3 & 0.535 & 56.0 & 577\\
Immortal \cite{wang2021immortal} & test & 67.7 & 0.599 & 57.2 & 320\\
GNN-PMB \cite{liu2022gnn} & test & 67.8 & 0.560 & 56.3 & 431\\
% NEBP \cite{liang2022neural} & test & 68.3 & 0.624 & \textbf{58.4} & 299\\
TransFusion-L \cite{bai2022transfusion} & test & 68.6 & \textbf{0.529} & 57.1 & 893\\
% Minkowski \cite{gwak2022minkowski} & test & 69.8 & 0.540 & 57.8 & 325\\
\cellcolor{Gray}\modelname & \cellcolor{Gray}test & \cellcolor{Gray}\textbf{70.1} & \cellcolor{Gray}0.549 & \cellcolor{Gray}\textbf{58.0} & \cellcolor{Gray}488\\

\bottomrule
\end{tabular}

%% file: 7conclusion.tex
\label{sec:conclusion}
We present \modelname, a simple and unified tracking framework designed to solve the problem of 2D and 3D MOT. \modelname incorporates object detection, motion prediction, and a detection-driven hierarchical data association, making it a comprehensive solution to MOT. The hierarchical data association strategy leverages detection scores as a powerful prior to identify the correct objects among low-score detections, reducing the issue of missing detections and fragmented trajectories. Additionally, our integrated motion prediction strategy for 3D MOT effectively addresses the problems of abrupt motion and object lost.
\modelname achieves state-of-the-art performance on both 2D and 3D MOT benchmarks. Furthermore, it possesses strong generalization capabilities and can be easily combined with different 2D and 3D detectors without any learnable parameters. 
We believe this simple and unified tracking framework will be useful in real-world applications.